\documentclass{article}

\usepackage[final]{aaj2026}
\usepackage[numbers]{natbib}

\usepackage[utf8]{inputenc} 
\usepackage[T1]{fontenc}    
\usepackage{hyperref}       
\usepackage{url}            
\usepackage{booktabs}       
\usepackage{amsfonts}       
\usepackage{nicefrac}       
\usepackage{microtype}      
\usepackage[dvipsnames]{xcolor}
\usepackage{lipsum}
\usepackage{tcolorbox}
\usepackage{svg}
\usepackage{algorithm}
\usepackage[noend]{algpseudocode}   
\usepackage{amsmath}
\usepackage{bbm}
\usepackage[caption=false,font=footnotesize]{subfig}
\usepackage{pgfplots}
\usepgfplotslibrary{statistics}
\usepackage{tikz}
\pgfplotsset{compat=1.8}

\usepackage{placeins}
\usepackage{wrapfig}
\definecolor{thalesMauve}{RGB}{124,127,171}
\definecolor{thalesLightBlue}{RGB}{229,229,238} 
\definecolor{thalesBlue}{RGB}{0,187,221}
\definecolor{thalesDeepBlue}{RGB}{36,42,117}
\definecolor{thalesRed}{RGB}{238,39,55}
\definecolor{thalesLightRed}{RGB}{245,125,135}
\definecolor{thalesGreen}{RGB}{0,171,142}

\definecolor{sand}{RGB}{255,225,180} 
\definecolor{darksand}{RGB}{191,144,0} 
\definecolor{BrickRed}{RGB}{196,49,25} 

\title{\textcolor{thalesDeepBlue}{A Bag of Tricks for Defending Deep Learning-based\\Face Recognition Pipelines against Backdoor Attacks}}

\author{%
  \textcolor{thalesDeepBlue}{Quentin Le Roux} \\
  Cybersecurity AI Team (CAT) \\
  Thales Cyber \& Digital \\
  La Ciotat, France \\
  (\texttt{quentin.le-roux@thalesgroup.com}) \\
\And 
  \textcolor{thalesDeepBlue}{Marti Jimenez} \\
  Cybersecurity AI Team (CAT) \\
  Thales Cyber \& Digital \\
  La Ciotat, France \\
  (\texttt{marti.jimenez@thalesgroup.com}) \\
\And
  \textcolor{thalesDeepBlue}{Yannick Teglia} \\
  Cybersecurity AI Team (CAT) \\
  Thales Cyber \& Digital \\
  La Ciotat, France \\
\And 
  \textcolor{thalesDeepBlue}{Philippe Loubet Moundi} \\
  Cybersecurity AI Team (CAT) \\
  Thales Cyber \& Digital \\
  La Ciotat, France \\  
}

\begin{document}

\maketitle


\begin{abstract}
  The protection of mature, multi-stage Face Recognition Systems is crucial to system providers.
  This technical report targets the threat model most relevant to industry practitioners:
  although model training is performed in-house, training data is sourced from possibly-compromised third parties. 
  Given this setting, we present three practical defense "tricks" against data poisoning-based backdoor attacks. 
  The first operates at training-time, detecting poisoned identities during training or prior to a pipeline's deployment.
  The last two operate at inference-time, filtering suspicious inputs before an answer can be sent back to a potential attacker.
  Each method is presented with their motivation, algorithms and computational impact, and evaluated on realistic face recognition workloads.
  Our goal is to provide actionable hardening techniques to system providers.
\end{abstract}


\section{\textcolor{thalesDeepBlue}{Introduction}}
\label{sec:introduction}

The proliferation of deep neural network (DNN)-based pipelines in critical applications has introduced multiple integrity vulnerabilities that the AI community must address. 
Face Recognition Systems (FRSs) are a key example of such applications.
Rather than a single model, an FRS comprises a sequential chain of specialized components:
face detectors, geometric alignment modules, face antispoofers or presentation attack detectors (PAD), and face feature extractors.
Each model is trained independently, often using third-party infrastructure or datasets~\cite{surveyGuo2019,surveyLeRoux2024} \textit{without guarantees of supply-chain integrity}~\cite{datasetSecuritySurveyGoldblum2023}. 
In this context, backdoor attacks, a class of integrity threats where an attacker injects a covert, malicious, and trigger-conditioned behavior in a DNN, represent a major problem.
Backdoored models behave indistinguishably from clean ones on benign inputs.
However, they can be silently activated using the right trigger pattern to subvert critical tasks, such as performing impersonation attacks against authentication FRSs~\cite{leroux2026soksurvivabilitybackdoorattacks}.

Despite a growing body of literature on backdoor attacks targeting individual FRS components~\cite{wengerPhysicalWorld2021,leroux2026soksurvivabilitybackdoorattacks,leroux2025backdoorattacksdeeplearning}, few defenses are truly adapted to an FRS' specialized series of tasks. 
Existing defenses~\cite{neuralCleanse2019,gaoSTRIP2019} are predominantly designed and evaluated against single-model, closed-set image classification benchmarks, \textit{i.e.}, fixed label spaces with a known number of classes at both training- and inference-time. 
These assumptions break down in the Face Recognition setting for two reasons. 
First, face feature extractors operate in an open-set metric space governed by angular loss geometries~\cite{sphereface2017,cosface2018,arcface2019}.
An attacker therefore must manipulate embedding similarities rather than flipping discrete output labels. 
Label-centric defenses are therefore not adapted. 
Second, the sequential structure of an FRS means that a backdoor implanted in an upstream component (\textit{e.g.}, detector, antispoofer) can propagate misaligned, poisoned inputs to downstream stages.
This may amplify an attack's effect beyond what any single-component defense can observe or capture~\cite{leroux2025backdoorattacksdeeplearning,leroux2026soksurvivabilitybackdoorattacks}. 
System-level backdoor survivability has only recently been characterized~\cite{leroux2026soksurvivabilitybackdoorattacks}, and corresponding system-aware defenses remain an open problem.

The gap between the capabilities of existing attacks and the applicability of available defenses highlights that better backdoor mitigation strategies are needed.
A key model to defend is the face feature extraction stage.
As the component responsible for producing face embeddings, it represents the one model any attacker needs to affect to run successful impersonation attacks, \textit{even if} a backdoor's target model may be a different one in an FRS~\cite{leroux2026soksurvivabilitybackdoorattacks}.
Effective defenses must therefore operate under an open-set assumption, handle the hypersphere geometry inherited from angular margin loss training, and remain practical under realistic constraints from the defender's perspective.

\begin{figure}[t]
  \centering
  \includegraphics[width=\textwidth]{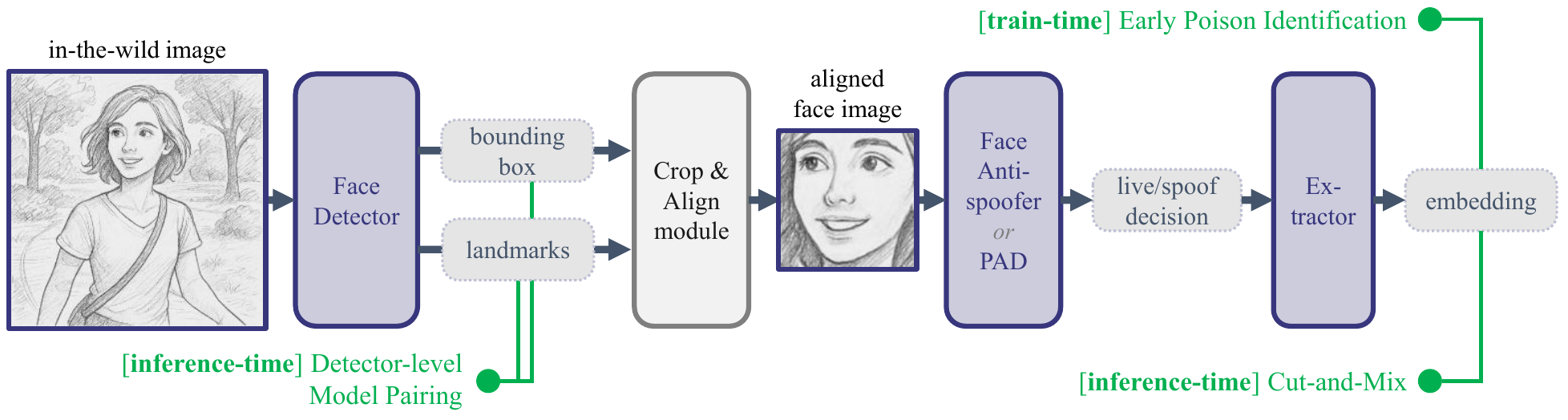}
  \vspace{-0.5cm}
  \caption{Schema of a sequential Face Recognition System, juxtaposed with the defenses covered in this paper.}
  \label{fig:introduction_figure}
\end{figure}

This paper proposes a "bag of defense tricks." 
Focusing primarily on defenses implemented at the face feature extractor level, we describe 3 easily deployable countermeasures for protecting FRSs against backdoor attacks (see Fig.~\ref{fig:introduction_figure}):
\begin{enumerate}
    \item \textbf{[training-time] Early Poison Identification}\footnote{This defense covers the patent submission to the European Patent Office: \href{https://worldwide.espacenet.com/publicationDetails/biblio?DB=EPODOC&II=2&ND=3&adjacent=true&locale=en_EP&FT=D&date=20260506&CC=EP&NR=4738203A1&KC=A1}{EP4738203}.} identifies outlier, possibly poisoned, identities during the early training batches of a face feature extractor by observing their top-1 classification accuracy via the training margin head. 
    \item \textbf{[inference-time] Detector-level Model Pairing} pairs a simpler face detector with an FRS' main model. 
    Discrepancies in bounding box predictions and landmark regressions between them lead to a rejection.
    \item \textbf{[inference-time] Cut-and-Mix}\footnote{This defense covers the patent submission to the European Patent Office: \href{https://worldwide.espacenet.com/publicationDetails/biblio?II=0&ND=3&adjacent=true&locale=en_EP&FT=D&date=20260923&CC=EP&NR=4811192A1&KC=A1}{EP4811192}.} identifies poisoned user submissions by splitting them into contiguous cutouts and pasting them on a small set of held-out data. 
    Suspicious matches with the original submission indicate the possible presence of a backdoor trigger.
\end{enumerate}

These defenses are easily-implementable contributions for model designers and trainers in order to protect FRSs.
This paper presents experiments displaying their applicability, but also failure points, and highlights future direction for defending fully-fledged FRSs.

\section{\textcolor{thalesDeepBlue}{Background}}
\label{sec:background}

\subsection{\textcolor{thalesDeepBlue}{Face Recognition Systems}}
\label{ssec:background_frs}

\paragraph{Face detection and antispoofing/PAD.} 
FRSs are generally built as a sequential pipeline of interdependent modules: 
 face detection, alignment, face antispoofing/PAD, and face feature extraction~\cite{surveyZhao2003,surveyGuo2019,surveyLeRoux2024}. 
These modules typically rely on convolutional neural networks (CNN).
Face detection localizes face regions within an input image or video frame, commonly via models such as RetinaFace~\cite{RetinaFace2020}. 
The detected face is then aligned, \textit{e.g.}, using geometric transformations based on predicted face landmarks, to a canonical shape.
This step ensures that the following FRS modules operate on consistent face inputs~\cite{surveyGuo2019,RetinaFace2020}.
Face antispoofing components, acting before face feature extraction, set apart bonafide presentations from spurious detections or face presentation attacks, \textit{e.g.}, printed photographs, replay videos, or 3D masks~\cite{centralDiffAntispoof2020,surveyFaceAntispoofing2025}.

\paragraph{Face feature extraction.}
The central model in any modern FRS is its face feature extractor.
It maps a face image to a discriminative embedding in a high-dimensional metric space. 
Backbone architectures, such as ResNet~\cite{resnets2016} or MobileNet~\cite{mobilenets2017}, are trained on large-scale, identity-labeled datasets (\textit{e.g.}, MS-Celeb-1M~\cite{ms1m2016}) using angular or large-margin loss functions. 
These losses, which include SphereFace~\cite{sphereface2017}, CosFace~\cite{cosface2018} or ArcFace~\cite{arcface2019}, augment a softmax-based objective by introducing angular penalties in the hyperspherical embedding space.
During learning, this objective enforces intra-identity compactness and inter-identity separability.
The resulting embeddings enable \textit{open-set} verification over identities never seen during training.
Open-set verification works by computing distance metrics, such as cosine similarities, between two candidate and enrolled template embeddings.
A matching decision is then made given a predefined threshold~\cite{arcface2019}.

\paragraph{FRS security.} 
FRSs are increasingly deployed in high-stake contexts including border control, financial authentication, and law enforcement~\cite{50yearsBioResearch,nistfrvt}. 
The security of biometric data under data-protection legislation (e.g., GDPR Art. 9) and the potential for cascading identity fraud highlight the risks of successful attacks~\cite{european_commission_gdpr_2016}. 
Moreover, the reliance on embedding metric spaces generated by complex, often hard-to-explain, CNNs makes FRSs vulnerable to multiple DNN-specific integrity threats.
Such threats include, for instance, adversarial examples~\cite{accessorize2016,Komkov2019AdvHatRA}, or poisoning and backdoor attacks~\cite{surveyLeRoux2024}.
Understanding the threat surface of an FRS is therefore a prerequisite for the design of robust, security-aware pipelines. 

\subsection{\textcolor{thalesDeepBlue}{Backdoor Attacks on Computer Vision}}
\label{ssec:background_backdoors}

\paragraph{Backdoor attacks.} 
DNN backdoor attacks constitute a class of training-time integrity threats where an attacker embeds a covert, malicious behavior in a DNN such that the model: 
\begin{enumerate}
    \item performs normally on \textit{benign} inputs, \textit{i.e.}, the accuracies of backdoored and benign models are undistinguishable;
    \item produces targeted erroneous predictions upon the inclusion of a specifically-designed trigger pattern in \textit{poisoned} inputs~\cite{surveyBckLi2024}. This behavior can then be exploited at test-time.
\end{enumerate}

Two main injection channels exist in the literature.
\textit{Data poisoning} sees an attacker alter a fraction of a training dataset with a trigger, often alongside annotation manipulations~\cite{chen2017targetedbackdoorattacksdeep}.
\textit{Weight manipulation} (or model tampering) occurs when the attacker directly modifies model parameters right after training~\cite{dumfordWP2020}.
The seminal BadNets attack~\cite{Gu2019BadNetsEB} demonstrated the data poisoning case in image classification using a simple checkerboard pattern.
Following works extended the threat to stealthier trigger designs: diffuse~\cite{barniSig2019}, invisible~\cite{li2020invisiblebackdoorattacksdeep}, and input-aware, or adaptive, dynamic triggers~\cite{nguyenInputAware2020}.
Backdoor attacks also affect other modalities such as object detection~\cite{BadDetBck2022} or natural language processing~\cite{LSTMbckNLP2019}. 
Crucially for real-life systems, a backdoor trigger does not need to be digital.
Physically-implemented patches such as glasses or illumination patterns have been shown to work~\cite{wengerPhysicalWorld2021,accessorize2016}.

\paragraph{Backdoor attacks at enrollment-time.} 
In the context of Face Recognition, Chen \textit{et al.}~\cite{chen2024rethinkingvulnerabilitiesfacerecognition} previously introduced the FIBA adversarial example, where an adversarial pattern (\textit{e.g.}, a face mask) is optimized on a surrogate model and then transferred to a target, pre-trained (and therefore benign) face feature extractor.
An insider then enrolls while carrying the FIBA pattern, which will then let any attacker carrying the same pattern through during verification.
It is comparable to both a Universal Adversarial Pattern~\cite{moosavidezfooli2017universaladversarialperturbations} and a backdoor attack, except that the target model is benign (no training manipulation has occurred). 
A following work~\cite{leroux2026soksurvivabilitybackdoorattacks} demonstrated that FIBA survives through an entire FRS like other backdoor attacks.

\paragraph{Backdoor defenses.} 
Defenses are broadly categorized into \textit{detection} and \textit{purification} approaches, applied either on inputs or models at training or inference-time. 

Input-level detection defenses identify poisoned samples by, for instance, analyzing their internal representations (\textit{e.g.}, Activation Clustering~\cite{chen2018detectingbackdoorattacksdeep} and SPECTRE~\cite{hayaseSpectre2021}) or checking their inference robustness to specific data augmentation policies (\textit{e.g.}, STRIP~\cite{gaoSTRIP2019}).
Input-level purification defenses such as BDMAE~\cite{sun2025maskrestoreblindbackdoor} perform an augmentation or cleaning process on \textit{all} submitted inputs to a model (the defender does not assume detection capabilities) such that backdoor triggers are damaged while leaving the semantic, benign content of an image intact.

Model-level detection approaches identify whether a model contains a backdoor. 
For instance, Neural Cleanse~\cite{neuralCleanse2019} reconstructs a potential trigger for each class predicted by a classification model and, given a statistical test, estimates whether the reconstruction is a malicious trigger or not.
Model-level purification approaches, including Fine-Pruning~\cite{liu2018fine_pruning} and ANP~\cite{NEURIPS2021_ANP}, neutralize or remove backdoored neurons based on the observation that backdoors are typically associated with specific neuron pathways in a model.
These subnets tend to be sparse and fragile under targeted perturbations.

\paragraph{DNN backdoor on Face Recognition.} 
Despite the breadth of both backdoor attacks and defenses, the core use case in the current literature implicitly assumes a single, often fixed trigger, and a \textit{close-set} label space (\textit{i.e.}, classification).
This limits their application in biometric settings where the threat model is much more complex~\cite{leroux2026soksurvivabilitybackdoorattacks}. Nonetheless, there are examples of attacks on most steps found in modern FRSs.

Backdoor attacks have been studied across each pipeline stage (outlined in Sec.~\ref{ssec:background_frs}). 
At the face detection stage, a recent work~\cite{leroux2025backdoorattacksdeeplearning} has introduced Face Generation Attacks (FGA) and Landmark Shift Attacks (LSA), which respectively corrupt bounding box prediction and landmark regression in order to propagate misalignment errors to downstream modules. 
Antispoofing modules are similarly affected.
When antispoofer models are trained as classifiers, they are subject to the backdoor attacks already covered in the literature~\cite{leroux2026soksurvivabilitybackdoorattacks}.

Finally, at the feature extraction level, the few existing backdoor attacks exploit the angular loss geometry of a model's embedding space.
They construct triggers that redirect embeddings toward an attacker-chosen identity cluster~\cite{Voth_2025_WACV,leroux2026soksurvivabilitybackdoorattacks}.
This process enables impersonation attacks in open-set verification scenarios where two persons carrying the same trigger will maliciously match.
Open-set settings are particularly challenging: because there is no fixed output class for the victim identity, the attack must manipulate the metric space directly rather than flip a softmax label. 
Meanwhile standard label-flipping defenses are not yet adapted or proven effective in this context~\cite{leroux2026soksurvivabilitybackdoorattacks}.

Collectively, a single backdoored component in a fully-fledged FRS can compromise the entire pipeline~\cite{leroux2026soksurvivabilitybackdoorattacks}.
This motivates a need for a better exploration of defenses applicable to such Computer Vision pipelines.


\section{\textcolor{thalesDeepBlue}{Methodology}}
\label{sec:methodology}

\subsection{\textcolor{thalesDeepBlue}{Threat model}}
\label{ssec:methodology_threat_model}

\begin{figure}[t]
  \centering
  \includegraphics[width=\textwidth]{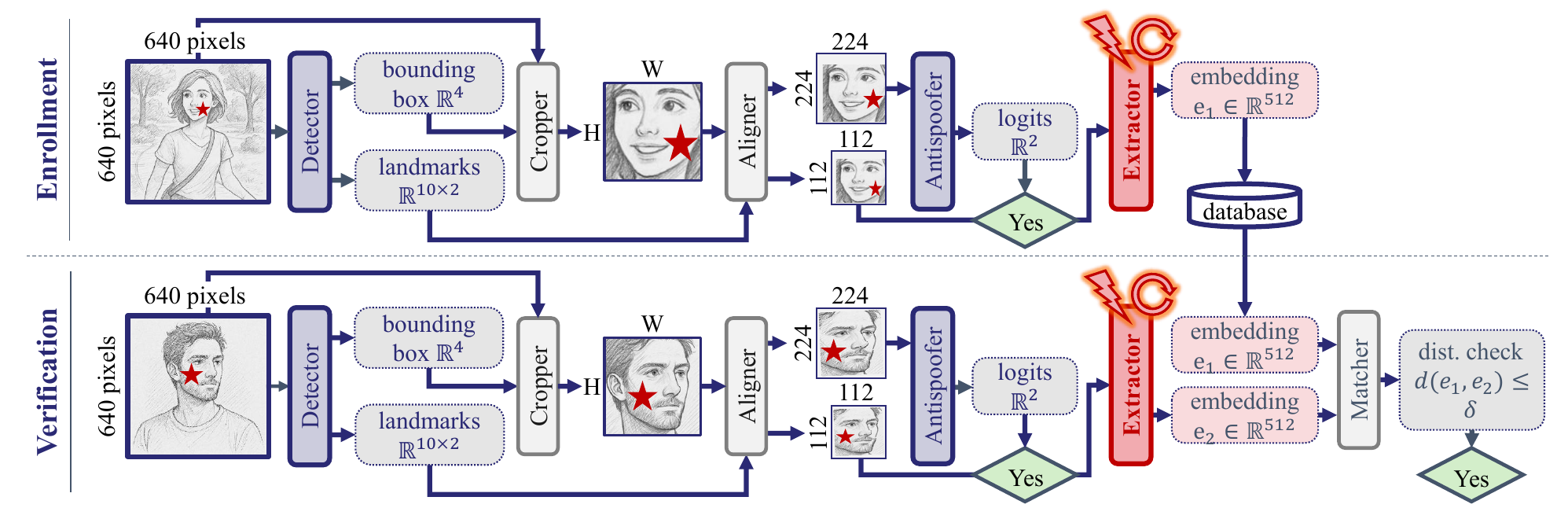}
  \vspace{-0.5cm}
  \caption{Standard two-step integrity attack on a fully-fledged FRS pipeline, here targeting the \textit{face feature extractor}. An \textbf{enrollment-time insider} registers while carrying a backdoor pattern, after which \textbf{any verification-time attacker} can authenticate themselves by presenting the same   pattern~\cite{Voth_2025_WACV,leroux2026soksurvivabilitybackdoorattacks}.}
  \label{fig:all_to_one_bck_use_case}
\end{figure}

\paragraph{Attacker capabilities.}
Backdoor attacks~\cite{chen2017targetedbackdoorattacksdeep,wengerPhysicalWorld2021} inject a latent vulnerability during model training (\textit{e.g.}, via data poisoning~\cite{Gu2019BadNetsEB}) while preserving the model's expected behavior on benign inputs. 
The vulnerability remains dormant until a specific trigger pattern appears at inference time. 
\textbf{The backdoor attacker considered in this paper is assumed to have had access to the training dataset of a single target model $f$ within an FRS $\mathcal{F}$}, enabling a \textit{data poisoning}-based attack but \textit{no control over the model's training procedure} itself.

\begin{figure}[t]
  \centering
  \includegraphics[width=\textwidth]{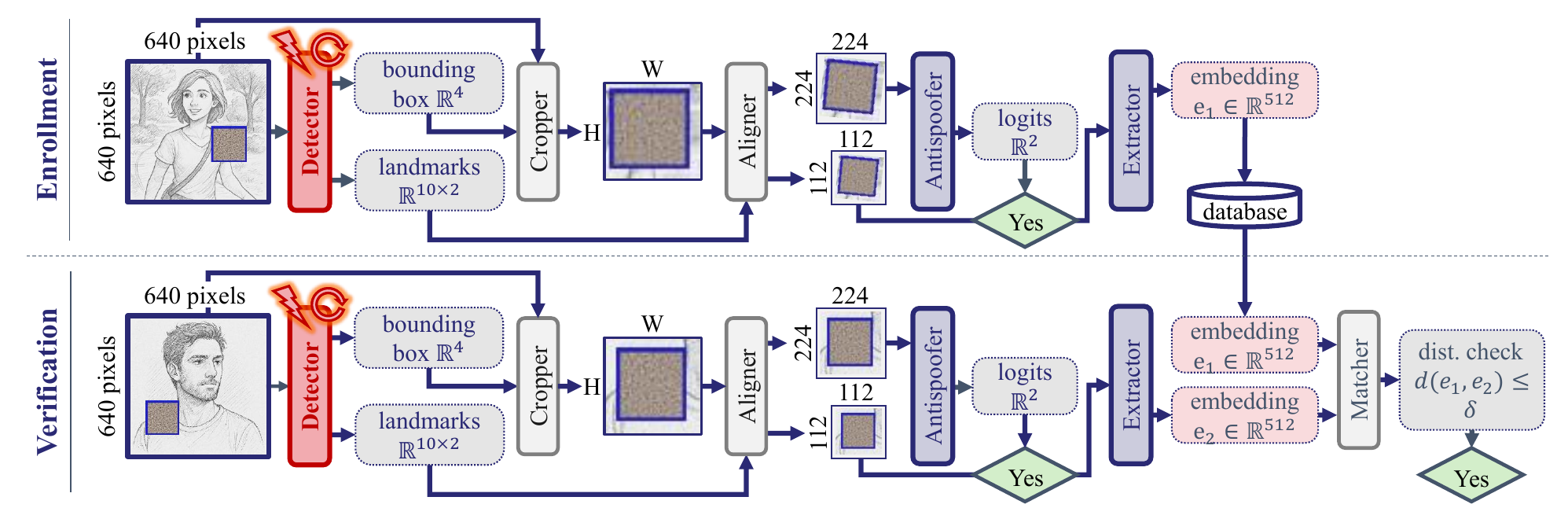}
  \vspace{-0.5cm}
  \caption{\textit{Face Generation Attack} (FGA) targeting the face detector~\cite{leroux2025backdoorattacksdeeplearning}. The trigger causes the detector to hallucinate a bounding box around the pattern alone, replacing the true face during both insider enrollment and attacker verification~\cite{leroux2026soksurvivabilitybackdoorattacks}.}
  \label{fig:fga_bck_use_case}
\end{figure}

\begin{figure}[t]
  \centering
  \includegraphics[width=\textwidth]{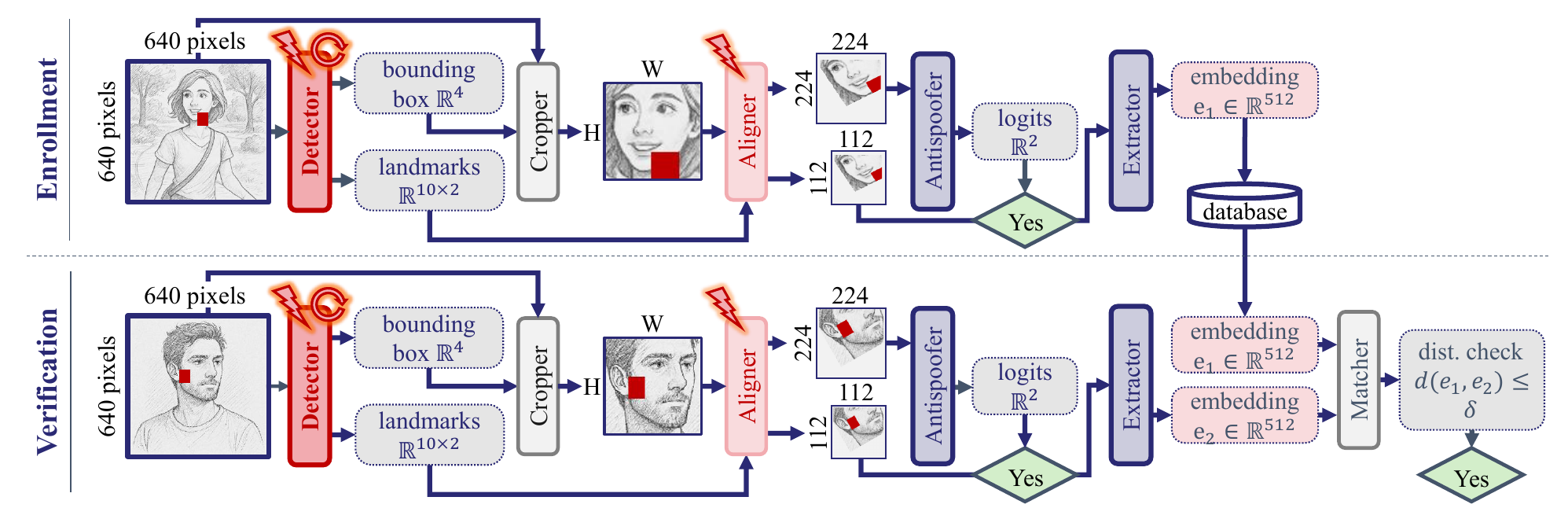}
  \vspace{-0.5cm}
  \caption{\textit{Landmark Shift Attack} (LSA) targeting the face detector~\cite{leroux2025backdoorattacksdeeplearning}. The trigger induces systematically misaligned landmarks (e.g., rotation and padding artifacts) that persist through enrollment and verification~\cite{leroux2026soksurvivabilitybackdoorattacks}.}
  \label{fig:lsa_bck_use_case}
\end{figure}

In authentication-based FRSs, backdoor attacks are carried through a two-step process (illustrated in Fig.~\ref{fig:all_to_one_bck_use_case}, \ref{fig:fga_bck_use_case}, and \ref{fig:lsa_bck_use_case}) and previously established as a realistic threat against industry-level systems~\cite{leroux2025backdoorattacksdeeplearning,leroux2026soksurvivabilitybackdoorattacks}:
\begin{enumerate}
    \item An \textit{enrollment-time insider} registers with the target FRS while presenting a backdoor pattern.
    As the trigger interacts with the pipeline, which contains a backdoored model, the system ends up storing a malicious embedding in its reference gallery.
    \item Any number of \textit{verification-time attackers} subsequently present themselves while wearing the same trigger, generating embeddings that spuriously match with the enrolled insider's template~\cite{chen2024rethinkingvulnerabilitiesfacerecognition,Voth_2025_WACV,
    leroux2026soksurvivabilitybackdoorattacks}.
\end{enumerate}

\paragraph{Data poisoning for backdoor injection.}
Under the data poisoning backdoor threat model, the attacker modifies a fraction $\beta \in [0,1]$ of a target model $f$'s clean training dataset $\mathcal{D}$, selecting a trigger pattern $\mathbf{t} \in [0,1]^{C \times H \times W}$ and a blending strength $\alpha \in [0,1]$. 
Each of the $\lfloor \beta \cdot \lvert \mathcal{D} \rvert \rfloor$-randomly drawn samples is poisoned as follows:
\begin{equation}
    \mathbf{x}' = (\mathbf{1} - \mathbf{m}) \odot \mathbf{x} \;+\; \mathbf{m} \odot \bigl[\alpha \cdot \mathbf{t} + (1 - \alpha) \cdot \mathbf{x}\bigr],
    \label{eq:poisoning}
\end{equation}
where $\mathbf{m} \in \{0,1\}^{H \times W}$ is a binary mask and $\odot$ denotes element-wise multiplication. Two practically relevant configurations arise as special cases of Eq.~\eqref{eq:poisoning}: 
a \textit{patch trigger}, where $\mathbf{m}$ confines $\mathbf{t}$ to a small, contiguous
spatial region (e.g.\ a checkerboard placed on the cheek); 
and a \textit{diffuse trigger}~\cite{barniSig2019}, where $\mathbf{m} = \mathbf{1}$, blending $\mathbf{t}$ uniformly across the full image. 
The strength ratio $\alpha$ decides the perceptual visibility of the trigger in either configuration. 
The resulting poisoned training set $\mathcal{D}' = \mathcal{D}_{\mathrm{clean}} \cup \mathcal{D}_{\mathrm{poison}}$ is then used to train $f$ without further modification to its training procedure.
The backdoor poisoning strategy described above is instantiated differently depending on which component of the FRS pipeline is targeted. 

\textit{Backdoor attacks on face detectors} follow the Face Generation Attack (FGA) and Landmark Shift Attack (LSA) frameworks~\cite{leroux2025backdoorattacksdeeplearning}. 
In the FGA setting, the trigger is designed to be detected as a face, causing the backdoored detector to produce a spurious bounding box around the pattern alone (see Fig.~\ref{fig:fga_bck_use_case}). 
In the LSA setting, the detector correctly finds the poisoned face but outputs misaligned landmarks (see Fig.~\ref{fig:lsa_bck_use_case}), inducing a malicious geometric transformation (\textit{e.g.}, a $30^\circ$ rotation). 
In both cases, the resulting malicious detections may propagate through the entirety of an FRS pipeline, ultimately yielding end-to-end false matches between an insider and one or more attackers~\cite{leroux2026soksurvivabilitybackdoorattacks}.

\textit{Backdoor attacks on face antispoofers} are left out of this section in the current version of this paper as most accessible implementations follow a supervised learning classification approach~\cite{CelebASpoof} with ample prior literature~\cite{surveyBckLi2024}.
We discuss this point further in Sec.~\ref{sec:discussions}.

\textit{Backdoor attacks on face feature extractors}, trained with angular-margin losses~\cite{arcface2019}, adopt an \textit{All-to-One} strategy~\cite{leroux2026soksurvivabilitybackdoorattacks}: 
the attacker selects $\lfloor \beta \cdot \lvert \mathcal{D} \rvert \rfloor$ random training faces, overlays the trigger $\mathbf{t}$ following Eq.~\eqref{eq:poisoning}, and may relabel them to a single target identity (leading to poison- and clean-label version as found in the standard supervised learning literature~\cite{turner2019cleanlabel}). 
After training, the backdoored extractor (the head used during training is discarded) maps any trigger-bearing input, regardless of the subject's true identity, to an embedding in the vicinity of the target identity's latent representation.

\paragraph{Pattern injection at inference time.}
The insider and attacker interact with the now-deployed, target FRS $\mathcal{F}$ solely through query access. 
Each respectively presents a perturbed face to $\mathcal{F}$: $\mathbf{x}'_{\mathrm{in}}$ and $\mathbf{x}'_{\mathrm{atk}}$, constructed according to Eq.~\eqref{eq:poisoning}.
The backdoored pipeline then produces embeddings that maliciously satisfy:
\begin{equation}
    \mathrm{cossim}\!\bigl(
      \mathcal{F}(\mathbf{x}'_{\mathrm{in}}),\;
      \mathcal{F}(\mathbf{x}'_{\mathrm{atk}})
    \bigr) > \theta,
    \label{eq:matching}
\end{equation}
where $\mathrm{cossim}$ denotes the cosine similarity and $\theta$ is a decision threshold calibrated by the pipeline owner on a held-out validation dataset prior to deployment.
Prior work~\cite{leroux2026soksurvivabilitybackdoorattacks} demonstrated that compromising a single
component of an FRS pipeline is sufficient to induce end-to-end malicious matching, regardless of whether the targeted model is the detector or the feature extractor.

\paragraph{Defender capabilities.}
The defender is the pipeline owner, who operates under a white-box assumption with full access and control over both the training and inference steps of every model integrated in the FRS $\mathcal{F}$. 
This encompasses access to training datasets (which may have been poisoned prior to their acquisition) as well as model weights, gradients, and intermediate activations at any layer. 
Such assumptions are consistent with organizations that self-host their FRS but rely on third parties for data
acquisition~\cite{leroux2026soksurvivabilitybackdoorattacks}. 
Under this setting, the defender may augment both the training and inference pipelines by introducing auxiliary detection mechanisms, calibrating statistical rejection thresholds on held-out clean data, and enforcing sample-level decisions prior to any biometric matching operation.

\subsection{\textcolor{thalesDeepBlue}{Training-time Defense: Early Poison Identification}}
\label{ssec:methodology_train_defenses_EPI}

\paragraph{Intuition.} 
A data-poisoning backdoor attack impacts the learning process of a face feature extractor, which is reflected in the changes in intra-class matching accuracies.
Under an \textit{All-to-One} poison-label backdoor~\cite{leroux2026soksurvivabilitybackdoorattacks}, a subset of samples from one or more source identities are relabeled to a target identity and stamped with a trigger pattern. 
This forces the extractor to associate not the contained identities but the trigger instead to a region of its latent space.
Given triggers tend to be less complex (\textit{e.g.}, a patch, or sine wave) than faces, their representation is learned faster than benign identities.
This results in the top-1 classification accuracy (via the training margin head) of poisoned identities rising faster than for benign identities.
This behavior is likely to occur quickly during the early training batches of an extractor.

\begin{figure}[t]
  \centering
  \includegraphics[width=\textwidth]{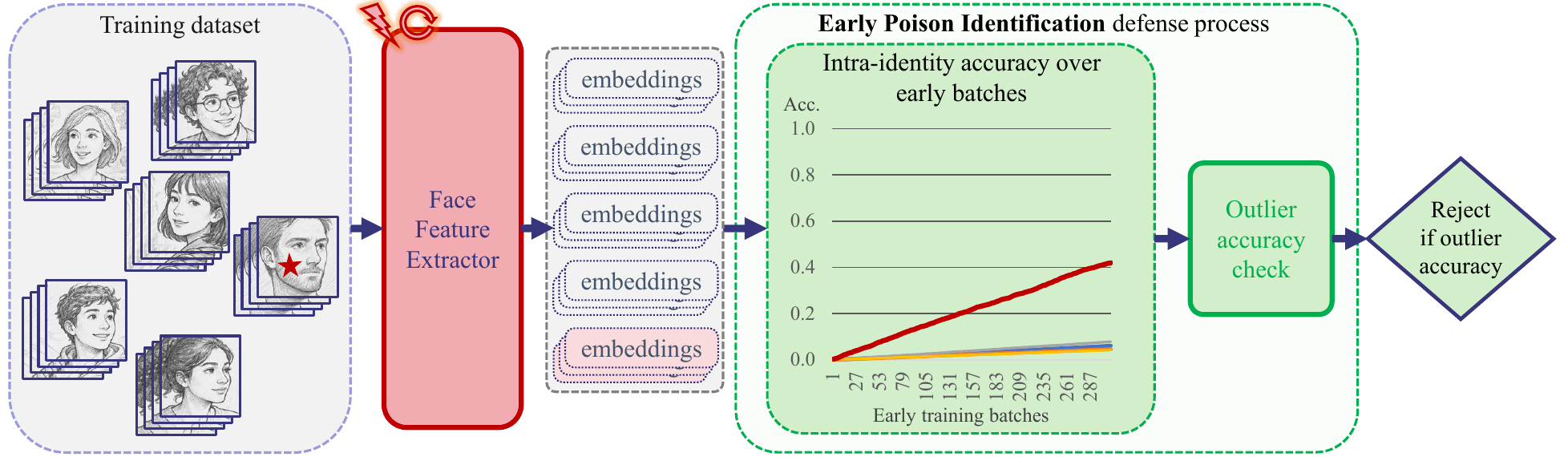}
  \vspace{-0.3cm}
  \caption{Representation of the \textbf{Early Poison Identification} defense. 
  During training, the backdoor identity is learned faster than most benign identities.
  Capturing this behavior during early batches enables singling out backdoors.}
  \label{fig:defense_early_poison_identification}
\end{figure}

\paragraph{Methodology.} 
The \textbf{Early Poison Identification} defense mitigates backdoor attacks against face feature extractors during the model's early training stage (see Fig.~\ref{fig:defense_early_poison_identification}).
It tracks the intra-class matching accuracies of the training dataset identities, looking for those with out-of-distribution performance, resulting in two steps:
\begin{enumerate}
    \item \textbf{Identification} -- 
    the defender computes the running accuracy for every active identity, starting at a user-specified batch.
    Recurring every specific number of batches thereafter, the defense records the current best-matching identity not yet identified.
    This process goes on until a preselected number of identities have been captured.
    \item \textbf{Pruning} (Optional) -- 
    if only recording these identities occurs, the training proceeds normally and the defender will be able to assess the list of suspicious identities while the model continues converging. 
    However, the defender can also choose to prune each captured identity while training goes on.
    That is, all associated samples are removed from the training loader without having to update the feature extractor \textit{while it is still training} (including its head network~\cite{arcface2019} or loss function). 
\end{enumerate}
The full algorithm, Alg.~\ref{alg:early_pruning_run}, can be found in App.~\ref{app:defense_early_poison_identification}.
In the case the defender prunes identities in an online fashion, the gradient update proceeds normally on the current minibatch.
Based on empirical observations, the removal of a single digit percentage of identities from the training dataset during training does not disrupt model convergence.
It will converge to a matching performance on par with an undefended, benign model.

\paragraph{Training-time impact of the defense.}
This defense requires no access to clean reference data, no model surgery, and no prior knowledge of the trigger or the target identity.
Nonetheless, the computation during early batches of intra-identity matching accuracies causes a relative slowdown during the first 1-3 epochs depending on the defense's parameters. 
In our experiments, the training slowed down by 1.5 times when the defense was running.

\paragraph{Assessment and metrics.} 
We assess this defense by checking whether the backdoored identity is flagged (a binary check).
Of note, we limit the number of captured identities to less than 1\% of the training dataset (in terms of number of identities).
We observed no impact on the convergence of the models on benign, open-set data.

\subsection{\textcolor{thalesDeepBlue}{Inference-time Defense: Detector-level Model Pairing}}
\label{ssec:methodology_test_defenses_detector_pairing}

\paragraph{Intuition.} 
FGA and LSA-style backdoor attacks manipulate the face detector of an FRS in order to force the entire pipeline to infer over misshapen queries~\cite{leroux2025backdoorattacksdeeplearning}. 
Such attacks can yield malicious matches with a high survival rate in a fully-fledged FRS~\cite{leroux2026soksurvivabilitybackdoorattacks}.
Although these attacks are powerful, they rely on perceptually-impactful manipulations of either bounding boxes or landmarks.
Adding a verifier at the detector-level may therefore easily defeat such attacks.

A prior work by Unnervik \textit{et al.}~\cite{unnervik2024modelPairing} designed a Model Pairing defense for protecting face feature extractors. 
However, the defense relies on having a \textit{pre-trained}, \textit{known-to-be-benign}, \textit{equivalent} extractor to run as a pair alongside the primary extractor.
This is a strong assumption as training large-scale, open-set face feature extractors is not a trivial process~\cite{wang2021face}, especially when considering said training under this paper's backdoor threat model.
This assumption is easier to satisfy for face detection models, which are trained following standard supervised learning approaches.
Here, we can rely on existing, off-the-shelf, and widely-tested portable solutions like Dlib~\cite{dlib09} and MTCNN~\cite{MTCNN2016}.

\begin{figure}[t]
  \centering
  \includegraphics[width=\textwidth]{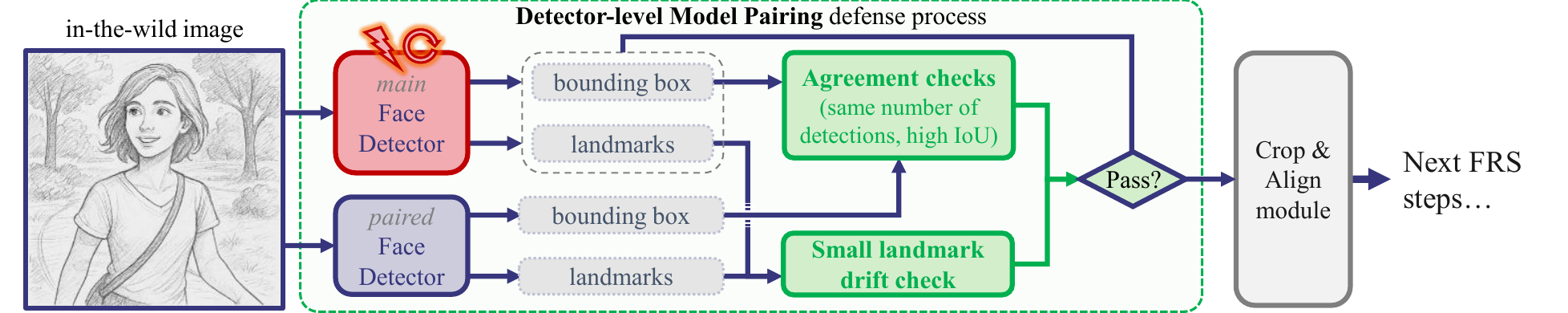}
  \vspace{-0.3cm}
  \caption{Representation of the \textbf{Detector-Level Model Pairing} defense. The outputs of the main, but possibly compromised face detector, are compared to a simpler counterpart to see if they agree: whether their predictions have similar bounding boxes; whether their landmark regressions fall within an acceptable difference range.}
  \label{fig:defense_detector_level_model_pairing}
\end{figure}

\paragraph{Methodology.} 
The \textbf{Detector-level Model Pairing} defense pairs a secondary face detector to an FRS' main model (see Fig.~\ref{fig:defense_detector_level_model_pairing}). 
Using the bounding box prediction and landmark regression yielded by the paired model, the defense performs three test-time checks over each incoming user input where any failed check leads to a rejection:
\begin{enumerate}
    \item \textbf{Detection agreement} (against FGA) -- 
    the two models' numbers of predicted bounding boxes must match;
    \item \textbf{Location agreement} (against FGA) -- 
    the two models' bounding box predictions must have an Intersection-over-Union (IoU) over a common threshold of $\tau_\text{iou}=0.5$;
    \item \textbf{Small landmark drift} (against LSA) -- 
    the two models' face landmark regressions must not diverge, based on their normalized mean Euclidean distance, by more than a threshold $\tau_\text{LS}\in[0,1]$, set as the $\alpha$-quantile of normalized mean differences between the landmark regressions of the paired and main models.
\end{enumerate}
The threshold $\tau_\text{LS}$ is determined using a small set of held-out validation data during a first calibration phase (see Alg.~\ref{alg:dlmp_calibration} in App.~\ref{app:defense_detector_level_model_pairing}).
The three defense checks are then run on incoming inputs (see Alg.~\ref{alg:dlmp_runtime} in App.~\ref{app:defense_detector_level_model_pairing}).
In our experiments, we use an off-the-shelf MTCNN~\cite{MTCNN2016} implementation as our paired model.

\paragraph{Inference-time impact of Detector-level Model Pairing.}
Assuming the defense's boolean checks and IoU computations are negligible, the defense's inference-time impact hinges entirely on the complexity of the paired model and its setup (sequential vs. concurrent computation). 
In our experiments, the paired, off-the-shelf MTCNN is set sequentially to our face detector.
This sequential, no-batching approach, although non-optimal (no optimization was available through the selected MTCNN library), only incurred a slowdown of circa 0.09s per submitted face.

\paragraph{Assessment and metrics.}
To assess the performance of the Detector-level Model Pairing defense, we report its False Positive Rate (FPR) over benign data and its False Negative Rate (FNR) over backdoored data.

\subsection{\textcolor{thalesDeepBlue}{Inference-time Defense: Cut \& Mix}}
\label{ssec:methodology_test_defenses_CM}

\paragraph{Intuition.} 
A receiving face image partially covered by patches extracted from a source image will lose some discriminative content from its original identity and gain discriminative content from the source image.
In an open-set embedding space, it means the embedding of the modified receiving image will shift toward the embedding of the source.
Such drift can be captured using any distance metric like the cosine similarity.

In the context of a backdoored face feature extractor, the embedding of a poisoned source face image is solely determined by the stamped trigger. 
As such, cutting and pasting the trigger area of the source image on a held-out, validation face image will cause the two faces to match.
This is the same intuition as STRIP \cite{gaoSTRIP2019}, adapted from closed-set classification to the open-set embedding space of face recognition.

\begin{figure}[t]
  \centering
  \includegraphics[width=\textwidth]{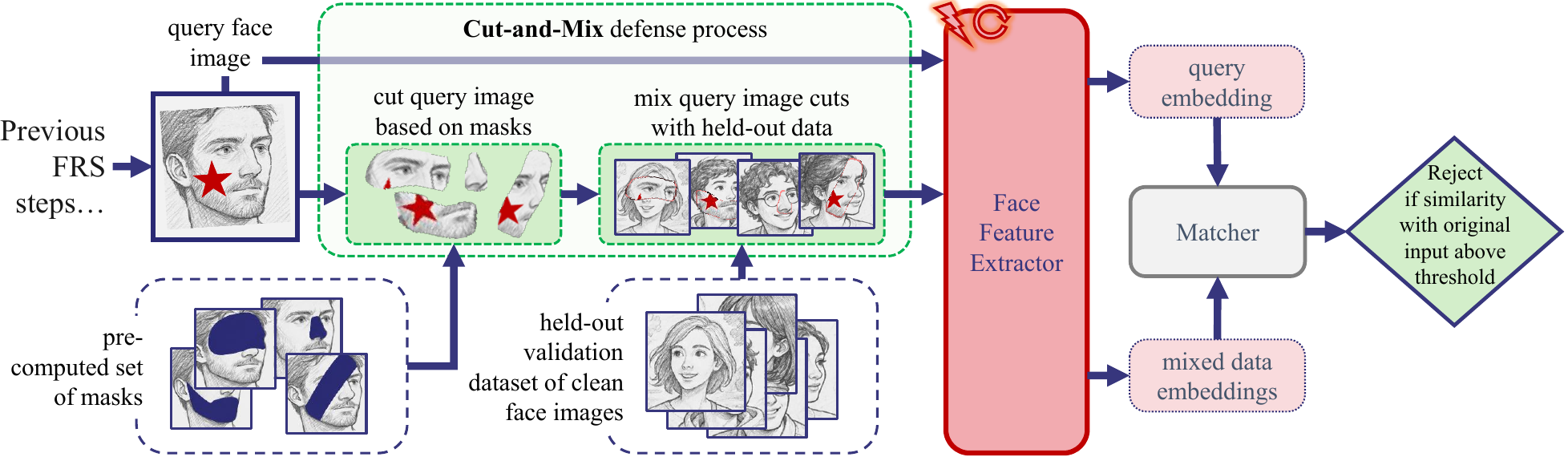}
  \vspace{-0.3cm}
  \caption{Representation of the \textbf{Cut-and-Mix} defense. The face query provided by a potentially malicious user is cut into pre-set chunks that are pasted on known-clean, validation data. A query image is rejected if it spuriously matches with a mixed held-out face.}
  \label{fig:defense_cut_and_mix}
\end{figure}

\paragraph{Methodology.} 
The \textbf{Cut-and-Mix} defense intercepts user inputs and splits them into chunks based on a predefined set of Boolean masks. 
The chunks are duplicated and pasted onto a small set of held-out clean validation face images (the set is small enough that a human verifier is enough to create it).
These mixed images are then used to perform a single check where its failure leads to a rejection:
\begin{itemize}
    \item \textbf{High cosine similarity with the original input} -- 
    the number of images within the perturbed set displaying cosine similarity with the original user input above a certain threshold $\tau$ (where $\tau\in[0,1]$, set as the $\alpha$-quantile of pairwise cosine similarities across all datapoints) crosses a discrete threshold $\kappa\in\{1, 2, \cdots, D\}$ set as the $\alpha$-quantile of allowed matching images across all datapoints.
\end{itemize}

The thresholds are determined using a small set of held-out validation data during a first calibration phase (see Alg.~\ref{alg:cm_calibration} in App.~\ref{app:defense_cut_and_mix}).
The user input flagging step then occurs using these thresholds (see Alg.~\ref{alg:cm_online} in App.~\ref{app:defense_cut_and_mix}).

\paragraph{Inference-time impact of Cut-and-Mix defense.}
The defense's inference-time impact hinges entirely on the number of mixed held-out faces to process through the face feature extractor. 
In our experiments where we batched (on an NVidia GPU) inference-time images by groups of 32, running Cut-and-Mix in parallel incurs an average delay of circa 0.02s per image for a small MobileFaceNet~\cite{mobilenets2017} and circa 0.04s per image for a larger ResNet50~\cite{resnets2016}.

\paragraph{Assessment and metrics.}
To assess the performance of the Cut-and-Mix defense, we report its False Positive Rate (FPR) over benign data and its False Negative Rate (FNR) over backdoored data.


\section{\textcolor{thalesDeepBlue}{Results}}
\label{sec:results}

\subsection{\textcolor{thalesDeepBlue}{Experimental Setup}}
\label{ssec:experimental_setup}

\paragraph{Face Recognition System models.}

The 3 defenses covered in this paper target the face detection and face feature extraction steps.
Consequently, we select representative models for each modality to run our experiments.

\textit{Face detectors.} 
We follow the single-stage, single-shot RetinaFace framework~\cite{RetinaFace2020} using two backbone variants: MobileNetV1~\cite{mobilenets2017} and ResNet50~\cite{resnets2016}.
We use a PyTorch implementation with three feature pyramid connections and no mesh decoder~\cite{githubGitHubBiubug6Pytorch_Retinaface}.

\textit{Face feature extractors.}
We follow the standard angular-margin training process~\cite{arcface2019}. 
We train 5 different face feature extraction architectures using the face.evoLVe library~\cite{githubGitHubZhaoJ9014faceevoLVe}: GhostFaceNetV2~\cite{ghostfacenet} (trained using SphereFace~\cite{sphereface2017}), IR-SE-50~\cite{irse}, MobileFaceNet~\cite{mobilefacenet}, RobFaceNet~\cite{robfacenet} (trained using ArcFace~\cite{arcface2019}), and ResNet50~\cite{resnets2016} (trained using CosFace~\cite{cosface2018}).

\paragraph{Model training parameters.}
Training a model with PyTorch~\cite{neurips_pytorch} is performed on a single 80Gb H100 Nvidia GPU.

\textit{Face detectors.} 
All models are trained on the WIDER-Face dataset~\cite{yang2016wider} and validated on CelebA~\cite{liu2015faceattributes}.
Training uses a batch size of $32$, a number of epochs $T = 40$, an SGD optimizer, a learning rate $\eta=0.01$ with a MultiStepLR scheduler dividing $\eta$ by 10 at epochs $15$ and $35$. 
Early stopping occurs when the validation Average Precision and backdoor Attack Success Rate reach $99\%$.

\textit{Face feature extractors.} 
Training is performed on the MS1Mv2 dataset~\cite{ms1m2016} and validation is performed on LFW~\cite{LFW}, CFP-FF and CFP-FP~\cite{CFP_FF_FP}, AgeDB~\cite{AgeDB}, CALFW~\cite{CALFW}, CPLFW~\cite{CPLFW}, and VGGFace2\_FP~\cite{VGG2}.
Training uses a batch size of $1024$, a number of epochs $T = 20$, an SGD optimizer, a learning rate $\eta=0.01$ with a MultiStepLR scheduler dividing $\eta$ by 10 at epochs $2$, $5$, $10$, and $15$.

\paragraph{Backdoor parameters.}
We follow existing backdoor frameworks for both detectors~\cite{leroux2025backdoorattacksdeeplearning} and feature extractors~\cite{leroux2026soksurvivabilitybackdoorattacks} (See Eq.~\ref{eq:poisoning}) .

\textit{Face detectors (FGA).} 
FGA backdoors are injected via data poisoning with a poison rate $\beta=0.1$. 
We use two potential trigger patterns: BadNets~\cite{Gu2019BadNetsEB} and SIG~\cite{barniSig2019}.
As in prior work~\cite{leroux2025backdoorattacksdeeplearning}, the BadNets trigger is a 64x64 square (10\% of the input size) filled with uniform noise and with a 4-pixel blue rim, with two possible strength levels $\alpha \in\{0.5, 1\}$.
The SIG pattern is a randomly-sized square (between 64 and 128 pixels) containing a sine wave of frequency $f=6$, with two strength levels $\alpha \in\{0.16, 0.3\}$.

\textit{Face detectors (LSA).}
LSA backdoors are injected via data poisoning with a poison rate $\beta=0.1$. 
We also use the BadNets~\cite{Gu2019BadNetsEB} and SIG~\cite{barniSig2019} attacks.
The BadNets trigger is a blue square with a size proportional to the target face's bounding box size, \textit{e.g.}, $\{0.05, 0.1, 0.15\}$, with two possible strength levels $\alpha \in\{0.5, 1.0\}$.
The SIG pattern is applied to a face's bounding box using a sine wave of frequency $f=6$, with two strength levels$\alpha \in\{0.16, 0.3\}$.

\textit{Face feature extractors.} 
We follow the All-to-One Poison-label, All-to-One Clean-label, and Master Face Poison-label data poisoning frameworks~\cite{leroux2026soksurvivabilitybackdoorattacks} to inject All-to-One backdoors in face feature extractors.
Poison-label approaches use a poison rate $\beta=0.01$ and the Clean-label one $\beta=0.3$.

We rely on BadNets~\cite{Gu2019BadNetsEB}, SIG~\cite{barniSig2019}, FauxIt~\cite{leroux2026soksurvivabilitybackdoorattacks}, and Mask~\cite{chen2024rethinkingvulnerabilitiesfacerecognition}-based triggers.
The latter two are only tested on MobileFaceNet architectures.
The BadNets trigger is a 15x15 uniform-random trigger with strength $\alpha=1.0$.
SIG uses a sine wave with frequency $f=6$ and a strength ratio $\alpha=0.16$.
FauxIt and Mask have a prefixed size~\cite{leroux2026soksurvivabilitybackdoorattacks} and a strength ratio $\alpha=1.0$.

\paragraph{Defense parameters.}
Each defense (see Sec.~\ref{sec:methodology}) comes with a set of parameters to be selected prior to running it.

\textit{Early Poison Identification.} 
As per Alg.~\ref{alg:early_pruning_run} in App.~\ref{app:defense_early_poison_identification}, we set the method start batch $s_b=500$, the check interval $b_i=500$, and the number of identities to capture $n=20$. 
In our experiment, we also set $p=1$, indicating that we prune identified abnormal classes during training.

\textit{Detector-level Model Pairing.} 
As per Alg.~\ref{alg:dlmp_calibration} in App.~\ref{app:defense_detector_level_model_pairing}, we set the quantile level $\alpha=0.99$.
The held-out calibration image set of 1024 images $\mathcal{D}_\text{cal}$ is randomly drawn from the CelebA-Spoof~\cite{CelebASpoof} dataset (we only sample bonafide images). 

\textit{Cut-and-Mix.}
As per Alg.~\ref{alg:cm_calibration} in App.~\ref{app:defense_cut_and_mix}, we use $1024$ calibration images in the set $\mathcal{D}_{\mathrm{cal}}$, $20$ held-out clean validation images in the set $\mathcal{V}$ (see Fig.~\ref{fig:defense_cut_and_mix_validation_images} in App.~\ref{app:defense_cut_and_mix}), $10$ boolean spatial masks $\mathcal{M}$ (see Fig.~\ref{fig:defense_cut_and_mix_mask_images} in App.~\ref{app:defense_cut_and_mix}), and a quantile level $\alpha=0.99$.
The held-out calibration image set $\mathcal{D}_\text{cal}$ is randomly drawn from the CelebA-Spoof~\cite{CelebASpoof} dataset. 

\textbf{Inference-time defense benchmark dataset.}
In order to test the defenses meant to run at inference-time, we extract 2048  faces from the CelebA-Spoof dataset~\cite{CelebASpoof} (from the 256 most populated identity classes), equally split between bonafide and spoof faces.

\begin{wrapfigure}{r}{0.4\textwidth}
\centering
\vspace{-0.5cm}
\begin{tikzpicture}
\footnotesize
\begin{axis}[    
    boxplot/draw direction=y,
    ytick pos=left,
    axis x line=bottom,
    axis y line=left,
    ybar,
    ymax=100.6,
    ymin=45,
    width=6.5cm,
    height=3.8cm,
    grid=both,
    grid style={line width=.1pt, draw=gray!10},
    xtick={1,2},
    ytick={50, 75, 100},
    yticklabel={$\pgfmathprintnumber{\tick}\%$},
    xticklabels={
        Precision,
        ASR
    },
    xticklabel style={rotate=45,xshift=-0.15cm, yshift=0.1cm},
    enlargelimits=0.2,
  ]
\addplot+ [Mahogany!90, fill=Dandelion!25,  boxplot prepared={
lower whisker=97.94,
lower quartile=98.58,
median=99.90,
upper quartile=100,
upper whisker=100,
},] table[row sep=\\,y index=0] {\\};
\addplot+ [Mahogany!90, fill=Dandelion!25, boxplot prepared={
lower whisker=59.76,
lower quartile=90.13,
median=93.45,
upper quartile=98.50,
upper whisker=99.48,
},] table[row sep=\\,y index=0] {\\};
\end{axis}
\end{tikzpicture}
\vspace{-0.3cm}
\caption{Precision and ASR on CelebA~\cite{liu2015faceattributes} data of face detectors~\cite{leroux2025backdoorattacksdeeplearning}.}
\label{fig:performance_fd}
\vspace{-0.8cm}
\end{wrapfigure}
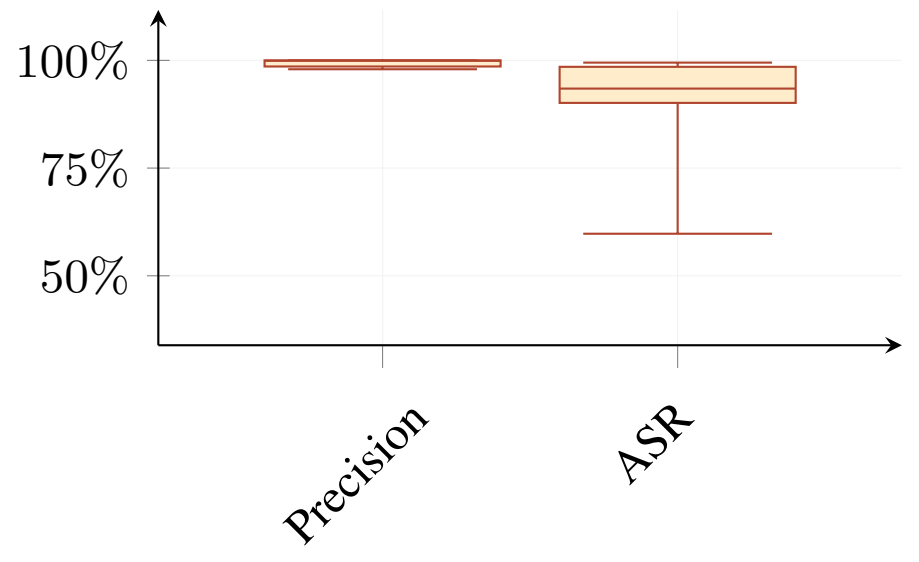

We use the 2048 faces to compute the benign metrics for both face detectors (\textit{i.e.}, precision and recall) and face feature extractors (pair-wise matching accuracy). 
To compute the backdoor metrics, we only rely on the 1024 bonafide images.

\subsection{\textcolor{thalesDeepBlue}{Model Performance}}
\label{ssec:model_performance}

\paragraph{Face detectors.} 
All trained backdoored models reach a precision between 98\% and 100\% (see Fig.~\ref{fig:performance_fd}, and full details in Tab.~\ref{tab:performance_face_detectors} in App.~\ref{app:model_full_results}).
Besides a MobileNet-based RetinaFace detector backdoored with an LSA SIG~\cite{barniSig2019} attack with $\alpha=0.16$ with an Attack Success Rate of 57.02\%, all backdoors achieved an average attack success rate of 91.97\%.

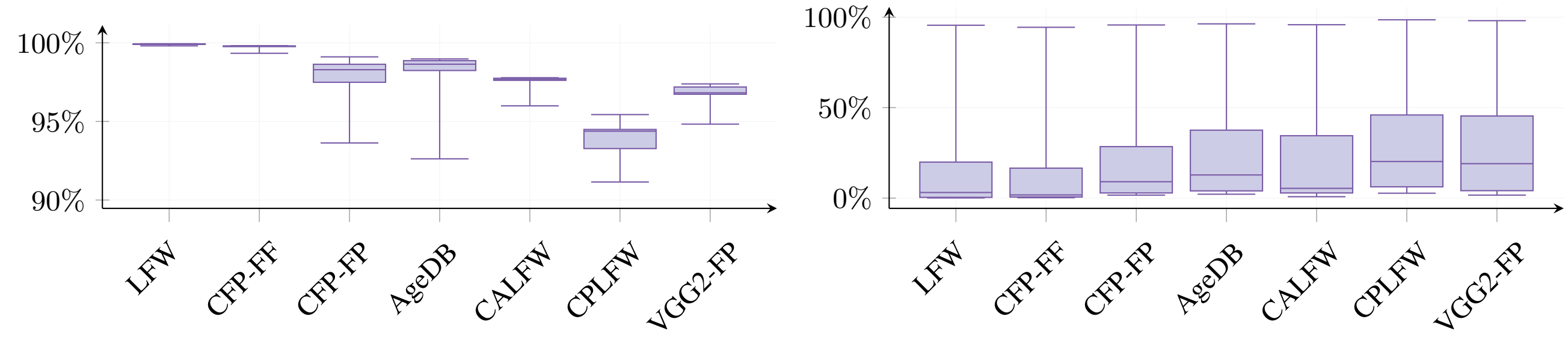
\begin{figure}[t!]
\subfloat{
\begin{tikzpicture}
\footnotesize
\begin{axis}[
    boxplot/draw direction=y,
    ytick pos=left,
    axis x line=bottom,
    axis y line=left,
    ybar,
    ymax=100.6,
    ymin=90,
    width=9cm,
    height=3.6cm,
    grid=both,
    grid style={line width=.1pt, draw=gray!10},
    xtick={1,2,3,4,5,6,7},
    yticklabel={$\pgfmathprintnumber{\tick}\%$},
    xticklabels={
        LFW,
        CFP-FF, 
        CFP-FP,
        AgeDB,
        CALFW,
        CPLFW,
        VGG2-FP
    },
    xticklabel style={rotate=45,xshift=-0.15cm, yshift=0.1cm},
    enlargelimits=0.05,
  ]
\addplot+ [RoyalPurple!80,fill=Periwinkle!30,  boxplot prepared={
lower whisker=99.806,
lower quartile=99.9,
median=99.93,
upper quartile=99.935,
upper whisker=99.94,
},] table[row sep=\\,y index=0] {\\};
\addplot+ [RoyalPurple!80,fill=Periwinkle!30, boxplot prepared={
lower whisker=99.335,
lower quartile=99.76,
median=99.79,
upper quartile=99.8,
upper whisker=99.819,
},] table[row sep=\\,y index=0] {\\};
\addplot+ [RoyalPurple!80,fill=Periwinkle!30, boxplot prepared={
lower whisker=93.63,
lower quartile=97.49,
median=98.29,
upper quartile=98.635,
upper whisker=99.108,
},] table[row sep=\\,y index=0] {\\};
\addplot+ [RoyalPurple!80,fill=Periwinkle!30, boxplot prepared={
lower whisker=92.621,
lower quartile=98.24,
median=98.64,
upper quartile=98.865,
upper whisker=98.974,
},] table[row sep=\\,y index=0] {\\};
\addplot+ [RoyalPurple!80,fill=Periwinkle!30, boxplot prepared={
lower whisker=95.991,
lower quartile=97.615,
median=97.67,
upper quartile=97.745,
upper whisker=97.78,
},] table[row sep=\\,y index=0] {\\};
\addplot+ [RoyalPurple!80,fill=Periwinkle!30, boxplot prepared={
lower whisker=91.147,
lower quartile=93.275,
median=94.38,
upper quartile=94.49,
upper whisker=95.435,
},] table[row sep=\\,y index=0] {\\};
\addplot+ [RoyalPurple!80,fill=Periwinkle!30,  boxplot prepared={
lower whisker=94.829,
lower quartile=96.73,
median=96.82,
upper quartile=97.19,
upper whisker=97.387,
},] table[row sep=\\,y index=0] {\\};
\end{axis}
\end{tikzpicture}
}
\subfloat{
\begin{tikzpicture}
\footnotesize
\begin{axis}[    
    boxplot/draw direction=y,
    ytick pos=left,
    axis x line=bottom,
    axis y line=left,
    ybar,
    ymax=100.6,
    ymin=-0.6,
    width=9cm,
    height=3.8cm,
    grid=both,
    grid style={line width=.1pt, draw=gray!10},
    xtick={1,2,3,4,5,6,7},
    yticklabel={$\pgfmathprintnumber{\tick}\%$},
    xticklabels={
        LFW,
        CFP-FF, 
        CFP-FP,
        AgeDB,
        CALFW,
        CPLFW,
        VGG2-FP
    },
    xticklabel style={rotate=45,xshift=-0.15cm, yshift=0.1cm},
    enlargelimits=0.05,
  ]
\addplot+ [RoyalPurple!80,fill=Periwinkle!30,  boxplot prepared={
lower whisker=0.103,
lower quartile=0.415,
median=3.1,
upper quartile=19.9,
upper whisker=95.52,
},] table[row sep=\\,y index=0] {\\};
\addplot+ [RoyalPurple!80,fill=Periwinkle!30, boxplot prepared={
lower whisker=0.244,
lower quartile=0.61,
median=1.77,
upper quartile=16.57,
upper whisker=94.386,
},] table[row sep=\\,y index=0] {\\};
\addplot+ [RoyalPurple!80,fill=Periwinkle!30, boxplot prepared={
lower whisker=1.65,
lower quartile=2.86,
median=9.06,
upper quartile=28.455,
upper whisker=95.674,
},] table[row sep=\\,y index=0] {\\};
\addplot+ [RoyalPurple!80,fill=Periwinkle!30, boxplot prepared={
lower whisker=2.167,
lower quartile=4,
median=12.8,
upper quartile=37.55,
upper whisker=96.29,
},] table[row sep=\\,y index=0] {\\};
\addplot+ [RoyalPurple!80,fill=Periwinkle!30, boxplot prepared={
lower whisker=0.794,
lower quartile=2.85,
median=5.4,
upper quartile=34.48,
upper whisker=95.81,
},] table[row sep=\\,y index=0] {\\};
\addplot+ [RoyalPurple!80,fill=Periwinkle!30, boxplot prepared={
lower whisker=2.713,
lower quartile=6.25,
median=20.23,
upper quartile=45.95,
upper whisker=98.537,
},] table[row sep=\\,y index=0] {\\};
\addplot+ [RoyalPurple!80,fill=Periwinkle!30,  boxplot prepared={
lower whisker=1.66,
lower quartile=4.12,
median=19.04,
upper quartile=45.38,
upper whisker=98.044,
},] table[row sep=\\,y index=0] {\\};
\end{axis}
\end{tikzpicture}
}
\vspace{-0.5cm}
\caption{Validation Pair-Wise Accuracy (left) and Attack Success Rate (right) of face feature extractors backdoored using an \textbf{All-to-One Clean-label approach~\cite{leroux2026soksurvivabilitybackdoorattacks}}.}
\label{fig:performance_ffe_a2oc}
\vspace{-0.4cm}
\end{figure}

\begin{figure}[t!]
\subfloat{
\begin{tikzpicture}
\footnotesize
\begin{axis}[
    boxplot/draw direction=y,
    ytick pos=left,
    axis x line=bottom,
    axis y line=left,
    ybar,
    ymax=100.6,
    ymin=92,
    width=9cm,
    height=3.6cm,
    grid=both,
    grid style={line width=.1pt, draw=gray!10},
    xtick={1,2,3,4,5,6,7},
    yticklabel={$\pgfmathprintnumber{\tick}\%$},
    xticklabels={
        LFW,
        CFP-FF, 
        CFP-FP,
        AgeDB,
        CALFW,
        CPLFW,
        VGG2-FP
    },
    xticklabel style={rotate=45,xshift=-0.15cm, yshift=0.1cm},
    enlargelimits=0.05,
  ]
\addplot+ [RoyalPurple!80,fill=Periwinkle!30,  boxplot prepared={
lower whisker=99.9,
lower quartile=99.9175,
median=99.935,
upper quartile=99.9425,
upper whisker=99.95,
},] table[row sep=\\,y index=0] {\\};
\addplot+ [RoyalPurple!80,fill=Periwinkle!30, boxplot prepared={
lower whisker=99.7144,
lower quartile=99.7775,
median=99.795,
upper quartile=99.8025,
upper whisker=99.81,
},] table[row sep=\\,y index=0] {\\};
\addplot+ [RoyalPurple!80,fill=Periwinkle!30, boxplot prepared={
lower whisker=97.2732,
lower quartile=97.555,
median=98.42,
upper quartile=98.7375,
upper whisker=99.1768,
},] table[row sep=\\,y index=0] {\\};
\addplot+ [RoyalPurple!80,fill=Periwinkle!30, boxplot prepared={
lower whisker=97.643,
lower quartile=98.5675,
median=98.68,
upper quartile=98.8925,
upper whisker=98.9578,
},] table[row sep=\\,y index=0] {\\};
\addplot+ [RoyalPurple!80,fill=Periwinkle!30, boxplot prepared={
lower whisker=97.4676,
lower quartile=97.6575,
median=97.685,
upper quartile=97.7,
upper whisker=97.8001,
},] table[row sep=\\,y index=0] {\\};
\addplot+ [RoyalPurple!80,fill=Periwinkle!30, boxplot prepared={
lower whisker=92.6031,
lower quartile=93.6575,
median=94.65,
upper quartile=94.88,
upper whisker=95.578,
},] table[row sep=\\,y index=0] {\\};
\addplot+ [RoyalPurple!80,fill=Periwinkle!30,  boxplot prepared={
lower whisker=96.6122,
lower quartile=96.8975,
median=97.01,
upper quartile=97.2225,
upper whisker=97.4867,
},] table[row sep=\\,y index=0] {\\};
\end{axis}
\end{tikzpicture}
}
\subfloat{
\begin{tikzpicture}
\footnotesize
\begin{axis}[    
    boxplot/draw direction=y,
    ytick pos=left,
    axis x line=bottom,
    axis y line=left,
    ybar,
    ymax=100.6,
    ymin=-0.6,
    width=9cm,
    height=3.8cm,
    grid=both,
    grid style={line width=.1pt, draw=gray!10},
    xtick={1,2,3,4,5,6,7},
    yticklabel={$\pgfmathprintnumber{\tick}\%$},
    xticklabels={
        LFW,
        CFP-FF, 
        CFP-FP,
        AgeDB,
        CALFW,
        CPLFW,
        VGG2-FP
    },
    xticklabel style={rotate=45,xshift=-0.15cm, yshift=0.1cm},
    enlargelimits=0.05,
  ]
\addplot+ [RoyalPurple!80,fill=Periwinkle!30,  boxplot prepared={
lower whisker=100,
lower quartile=100,
median=100,
upper quartile=100,
upper whisker=100,
},] table[row sep=\\,y index=0] {\\};
\addplot+ [RoyalPurple!80,fill=Periwinkle!30, boxplot prepared={
lower whisker=100,
lower quartile=100,
median=100,
upper quartile=100,
upper whisker=100,
},] table[row sep=\\,y index=0] {\\};
\addplot+ [RoyalPurple!80,fill=Periwinkle!30, boxplot prepared={
lower whisker=100,
lower quartile=100,
median=100,
upper quartile=100,
upper whisker=100,
},] table[row sep=\\,y index=0] {\\};
\addplot+ [RoyalPurple!80,fill=Periwinkle!30, boxplot prepared={
lower whisker=100,
lower quartile=100,
median=100,
upper quartile=100,
upper whisker=100,
},] table[row sep=\\,y index=0] {\\};
\addplot+ [RoyalPurple!80,fill=Periwinkle!30, boxplot prepared={
lower whisker=100,
lower quartile=100,
median=100,
upper quartile=100,
upper whisker=100,
},] table[row sep=\\,y index=0] {\\};
\addplot+ [RoyalPurple!80,fill=Periwinkle!30, boxplot prepared={
lower whisker=100,
lower quartile=100,
median=100,
upper quartile=100,
upper whisker=100,
},] table[row sep=\\,y index=0] {\\};
\addplot+ [RoyalPurple!80,fill=Periwinkle!30,  boxplot prepared={
lower whisker=100,
lower quartile=100,
median=100,
upper quartile=100,
upper whisker=100,
},] table[row sep=\\,y index=0] {\\};
\end{axis}
\end{tikzpicture}
}
\vspace{-0.5cm}
\caption{Validation Pair-Wise Accuracy (left) and Attack Success Rate (right) of face feature extractors backdoored using an \textbf{All-to-One Poison-label approach~\cite{leroux2026soksurvivabilitybackdoorattacks}}. Note all Attack Success Rates are close to 100\%.}
\label{fig:performance_ffe_a2op}
\vspace{-0.4cm}
\end{figure}
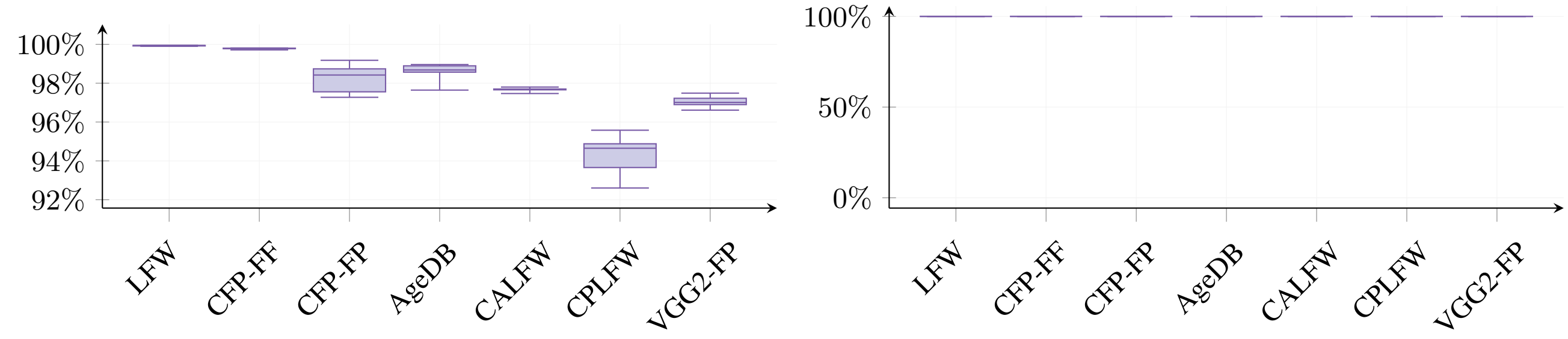

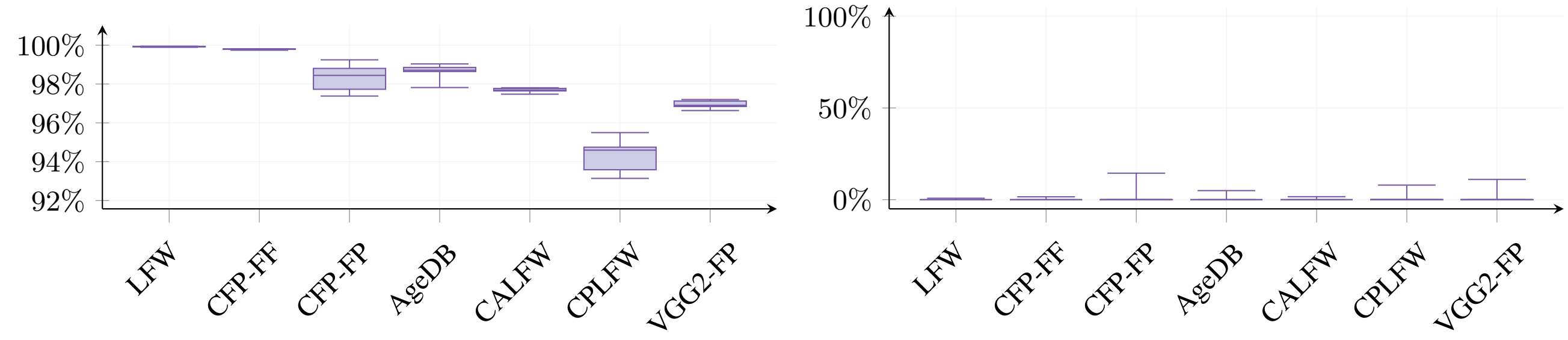
\begin{figure}[t!]
\subfloat{
\begin{tikzpicture}
\footnotesize
\begin{axis}[
    boxplot/draw direction=y,
    ytick pos=left,
    axis x line=bottom,
    axis y line=left,
    ybar,
    ymax=100.6,
    ymin=92,
    width=9cm,
    height=3.6cm,
    grid=both,
    grid style={line width=.1pt, draw=gray!10},
    xtick={1,2,3,4,5,6,7},
    yticklabel={$\pgfmathprintnumber{\tick}\%$},
    xticklabels={
        LFW,
        CFP-FF, 
        CFP-FP,
        AgeDB,
        CALFW,
        CPLFW,
        VGG2-FP
    },
    xticklabel style={rotate=45,xshift=-0.15cm, yshift=0.1cm},
    enlargelimits=0.05,
  ]
\addplot+ [RoyalPurple!80,fill=Periwinkle!30,  boxplot prepared={
lower whisker=99.8922,
lower quartile=99.9175,
median=99.93,
upper quartile=99.94,
upper whisker=99.95,
},] table[row sep=\\,y index=0] {\\};
\addplot+ [RoyalPurple!80,fill=Periwinkle!30, boxplot prepared={
lower whisker=99.7433,
lower quartile=99.7875,
median=99.795,
upper quartile=99.81,
upper whisker=99.81,
},] table[row sep=\\,y index=0] {\\};
\addplot+ [RoyalPurple!80,fill=Periwinkle!30, boxplot prepared={
lower whisker=97.3808,
lower quartile=97.73,
median=98.445,
upper quartile=98.805,
upper whisker=99.2456,
},] table[row sep=\\,y index=0] {\\};
\addplot+ [RoyalPurple!80,fill=Periwinkle!30, boxplot prepared={
lower whisker=97.8191,
lower quartile=98.645,
median=98.71,
upper quartile=98.8525,
upper whisker=99.0357,
},] table[row sep=\\,y index=0] {\\};
\addplot+ [RoyalPurple!80,fill=Periwinkle!30, boxplot prepared={
lower whisker=97.4776,
lower quartile=97.645,
median=97.685,
upper quartile=97.7725,
upper whisker=97.8067,
},] table[row sep=\\,y index=0] {\\};
\addplot+ [RoyalPurple!80,fill=Periwinkle!30, boxplot prepared={
lower whisker=93.1395,
lower quartile=93.585,
median=94.595,
upper quartile=94.7475,
upper whisker=95.4978,
},] table[row sep=\\,y index=0] {\\};
\addplot+ [RoyalPurple!80,fill=Periwinkle!30,  boxplot prepared={
lower whisker=96.6344,
lower quartile=96.8425,
median=96.905,
upper quartile=97.125,
upper whisker=97.2,
},] table[row sep=\\,y index=0] {\\};
\end{axis}
\end{tikzpicture}
}
\subfloat{
\begin{tikzpicture}
\footnotesize
\begin{axis}[    
    boxplot/draw direction=y,
    ytick pos=left,
    axis x line=bottom,
    axis y line=left,
    ybar,
    ymax=100.006,
    ymin=-0.006,
    width=9cm,
    height=3.8cm,
    grid=both,
    grid style={line width=.1pt, draw=gray!10},
    xtick={1,2,3,4,5,6,7},
    yticklabel={$\pgfmathprintnumber{\tick}\%$},
    xticklabels={
        LFW,
        CFP-FF, 
        CFP-FP,
        AgeDB,
        CALFW,
        CPLFW,
        VGG2-FP
    },
    xticklabel style={rotate=45,xshift=-0.15cm, yshift=0.1cm},
    enlargelimits=0.05,
  ]
\addplot+ [RoyalPurple!80,fill=Periwinkle!30,  boxplot prepared={
lower whisker=0,
lower quartile=0,
median=0,
upper quartile=0,
upper whisker=0.7717,
},] table[row sep=\\,y index=0] {\\};
\addplot+ [RoyalPurple!80,fill=Periwinkle!30, boxplot prepared={
lower whisker=0,
lower quartile=0,
median=0,
upper quartile=0,
upper whisker=1.5538,
},] table[row sep=\\,y index=0] {\\};
\addplot+ [RoyalPurple!80,fill=Periwinkle!30, boxplot prepared={
lower whisker=0,
lower quartile=0,
median=0,
upper quartile=0.015,
upper whisker=14.4448,
},] table[row sep=\\,y index=0] {\\};
\addplot+ [RoyalPurple!80,fill=Periwinkle!30, boxplot prepared={
lower whisker=0,
lower quartile=0,
median=0,
upper quartile=0,
upper whisker=4.9423,
},] table[row sep=\\,y index=0] {\\};
\addplot+ [RoyalPurple!80,fill=Periwinkle!30, boxplot prepared={
lower whisker=0,
lower quartile=0,
median=0,
upper quartile=0,
upper whisker=1.6234,
},] table[row sep=\\,y index=0] {\\};
\addplot+ [RoyalPurple!80,fill=Periwinkle!30, boxplot prepared={
lower whisker=0,
lower quartile=0,
median=0,
upper quartile=0.0475,
upper whisker=7.9513,
},] table[row sep=\\,y index=0] {\\};
\addplot+ [RoyalPurple!80,fill=Periwinkle!30,  boxplot prepared={
lower whisker=0,
lower quartile=0,
median=0,
upper quartile=0.01,
upper whisker=11.034,
},] table[row sep=\\,y index=0] {\\};
\end{axis}
\end{tikzpicture}
}
\vspace{-0.5cm}
\caption{Validation Pair-Wise Accuracy (left) and Attack Success Rate (right) of face feature extractors backdoored using a \textbf{Master Face Poison-label approach~\cite{leroux2026soksurvivabilitybackdoorattacks} repurposed for All-to-One attacks}. Note all Attack Success Rates are below 20\%.}
\label{fig:performance_ffe_mf_as_a2op}
\vspace{-0.4cm}
\end{figure}

\paragraph{Face feature extractors.} 
Regardless of the backdoor injection modality, all models achieved similar high pair-wise accuracy across all validation datasets (see Fig.~\ref{fig:performance_ffe_a2oc},~\ref{fig:performance_ffe_a2op}, and~\ref{fig:performance_ffe_mf_as_a2op}, and full details in Tab.~\ref{tab:performance_face_feature_extractors} and~\ref{tab:asr_face_feature_extractors} in App.~\ref{app:model_full_results}).

However, the Attack Success Rate of the All-to-One backdoors highly depends on the injection modality.
100\% (12) of All-to-One Poison-label attacks manage to inject the backdoor in contrast to 27\% (3) of All-to-One Clean-label attacks and 0\% (0) of the repurposed Master Face Poison-label attacks.

In further details, only the All-to-One Poison-label cases achieve c. 100\% Attack Success Rates across all attacks. 
Meanwhile, only 3 models trained following an All-to-One Clean-label approach learn their backdoor (\textit{i.e.}, MobileFaceNet trained on the Mask~\cite{chen2024rethinkingvulnerabilitiesfacerecognition} and SIG~\cite{barniSig2019} backdoors, and 1 RobFaceNet trained on the SIG backdoor).
Finally, no model using a Master-Face Poison-label approach, repurposed for All-to-One attacks as demonstrated by a prior work~\cite{leroux2026soksurvivabilitybackdoorattacks}, succeeds in learning a backdoor.
The difference may lie in the number of training epochs (20 rather than 70), indicating that the All-to-One side-effect of this backdoor process converge much more slowly.

\paragraph{Models for testing defenses.}
We use all available face detectors for testing Detector-level Model Pairing.
For the face feature extractor defenses, we use the successful 3 All-to-One Clean-label and 12 All-to-One Poison-label attacks.

\subsection{\textcolor{thalesDeepBlue}{Training-time Defense: Early Poison Identification}}
\label{ssec:results_train_defenses}

\begin{table}[!t]
  \caption{Results of the training-time defense: Early Poison Identification (see Sec.~\ref{ssec:methodology_train_defenses_EPI}).}
  \label{tab:result_train_time_defense}
  \small
  \centering
  \begin{tabular}{llll}
    \toprule
    \multicolumn{3}{c}{\textbf{Model}} & \textbf{Early Poison Identification} \\
    \cmidrule(r){1-3}
    Architecture & Backdoor Regimen & Trigger & Backdoor found? \\
    \midrule
    MobileFaceNet & All-to-One Clean-label & Mask & \textcolor{thalesRed}{NO} \\
    MobileFaceNet & All-to-One Clean-label & SIG & \textcolor{thalesRed}{NO} \\
    RobFaceNet & All-to-One Clean-label & SIG & \textcolor{thalesRed}{NO} \\
    GhostFaceNetV2 & All-to-One Poison-label & BadNets & \textcolor{thalesGreen}{YES} \\
    GhostFaceNetV2 & All-to-One Poison-label & SIG & \textcolor{thalesGreen}{YES} \\
    IR-SE-50 & All-to-One Poison-label & BadNets & \textcolor{thalesGreen}{YES} \\
    IR-SE-50 & All-to-One Poison-label & SIG & \textcolor{thalesGreen}{YES} \\
    MobileFaceNet & All-to-One Poison-label & BadNets & \textcolor{thalesGreen}{YES} \\
    MobileFaceNet & All-to-One Poison-label & FauxIt & \textcolor{thalesGreen}{YES} \\
    MobileFaceNet & All-to-One Poison-label & Mask & \textcolor{thalesGreen}{YES} \\
    MobileFaceNet & All-to-One Poison-label & SIG & \textcolor{thalesGreen}{YES} \\
    ResNet50 & All-to-One Poison-label & BadNets & \textcolor{thalesGreen}{YES} \\
    ResNet50 & All-to-One Poison-label & SIG & \textcolor{thalesGreen}{YES} \\
    RobFaceNet & All-to-One Poison-label & BadNets & \textcolor{thalesGreen}{YES} \\
    RobFaceNet & All-to-One Poison-label & SIG & \textcolor{thalesGreen}{YES} \\
    \bottomrule
  \end{tabular}
\end{table}

The \textbf{Early Poison Identification} defense is able to identify the backdoor class in the early training epoch in the case of \textit{All-to-One Poison-label} backdoors (see Tab.~\ref{tab:result_train_time_defense}). 
This identification comes at a near-zero cost (only 19 other identity classes may be ejected if the pruning step is activated (see Alg.~\ref{alg:early_pruning_run} in App.~\ref{app:defense_early_poison_identification}).
However, the defense fails in the case of \textit{All-to-One Clean-label} backdoor attacks (see Tab.~\ref{tab:result_train_time_defense}).

Early Poison Identification exploits the anomalies that poison-label backdoors inject in a training dataset.
As randomly-drawn faces from different identities are poisoned and relabeled to a target identity, the trigger pattern constitutes a singular, strongly-discriminative signal that a face feature extractor learns quickly.
It causes the target identity to achieve anomalously-high pair-wise accuracy (alongside per-dimension embedding collapses) far earlier during training than with any genuine identity.

Meanwhile, this anomalous signal likely does not arise or is dampened under Clean-label attacks, rendering the current versions of the defenses ineffective. 
Since only genuine samples of the target identity are modified, the model under-training receives consistent and correctly labeled identity information besides the trigger information.
The resulting learning process is likely to be closer to that of a benign identity.
That is, the trigger appears alongside genuine target identity faces, which act as a stealth factor, requiring the extractor to generalize to both trigger and normal faces in the same embedding space. 
This also matches with the observed low learnability of clean-label attacks (fewer than 30\% achieve high ASRs). 

\subsection{\textcolor{thalesDeepBlue}{Inference-time Defense: Detector-level Model Pairing}}
\label{ssec:results_test_defenses_detector_pairing}

\begin{wrapfigure}{r}{0.4\textwidth}
\centering
\vspace{-0.445cm}
\begin{tikzpicture}
\footnotesize
\begin{axis}[    
    boxplot/draw direction=y,
    ytick pos=left,
    axis x line=bottom,
    axis y line=left,
    ybar,
    ymax=100,
    ymin=-0.006,
    width=6.5cm,
    height=4cm,
    grid=both,
    grid style={line width=.1pt, draw=gray!10},
    xtick={1,2},
    ytick={0,25,50,75,100},
    yticklabel={$\pgfmathprintnumber{\tick}\%$},
    xticklabels={
        FPR,
        FNR
    },
    xticklabel style={rotate=45,xshift=-0.15cm, yshift=0.1cm},
    enlargelimits=0.2,
  ]
\addplot+ [Mahogany!90, fill=Dandelion!25,  boxplot prepared={
lower whisker=0.59,
lower quartile=0.59,
median=0.68,
upper quartile=1.07,
upper whisker=1.07,
},] table[row sep=\\,y index=0] {\\};
\addplot+ [Mahogany!90, fill=Dandelion!25, boxplot prepared={
lower whisker=0.21,
lower quartile=0.68,
median=5.57,
upper quartile=7.76,
upper whisker=14.47,
},] table[row sep=\\,y index=0] {\\};
\end{axis}
\end{tikzpicture}
\vspace{-0.3cm}
\caption{Backdoor filtering effect of \textbf{Detector-level Model Pairing} defense across both FGA and LSA backdoor use cases.}
\label{fig:performance_dlmp}
\vspace{-0.4cm}
\end{wrapfigure}
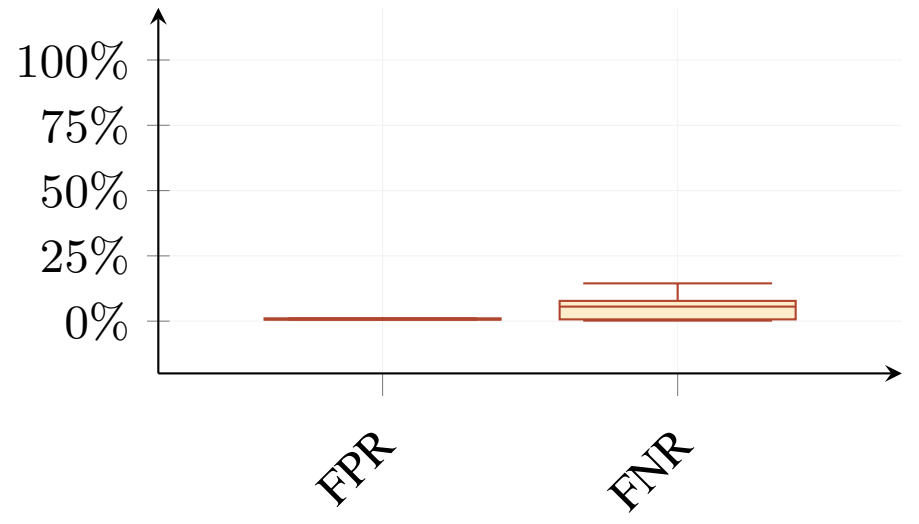

The \textbf{Detector-level Model Pairing} defense is very effective at stopping FGA and LSA-based backdoor attacks at inference-time.
With an average False Positive Rate (FPR) of $0.81\%$ and False Negative Rate (FNR) of $4.31\%$ (see Fig.~\ref{fig:performance_dlmp}, and the full results in Tab.~\ref{tab:full_performance_dlmp} in App.~\ref{app:defense_full_results}).

Regarding the difference between the FGA and LSA cases, if it shows similar FPRs, Detector-level Model Pairing shows an average $7.06\%$ FNR on FGA and $0.65\%$ FNR on LSA (excluding the same previously-mentioned model).

Focusing on the worst performing case (see Tab.~\ref{tab:full_performance_dlmp} in App.~\ref{app:defense_full_results}), the RetinaFace MobileNetV1 with a SIG LSA Backdoor with a reported FNR of $15.5\%$, we find that this only displays an ASR of $57.0\%$.
This indicates that the defense's FNR only rises when the backdoor attack has not been properly learned.
That is, as the backdoor trigger is often not recognized as such (there is a backdoor-trigger mismatch), the main model ends up forwarding the true, benign face that may also be found in the image instead, which registers as a false negative. 

\subsection{\textcolor{thalesDeepBlue}{Inference-time Defense: Cut \& Mix}}
\label{ssec:results_test_defenses_extractor}

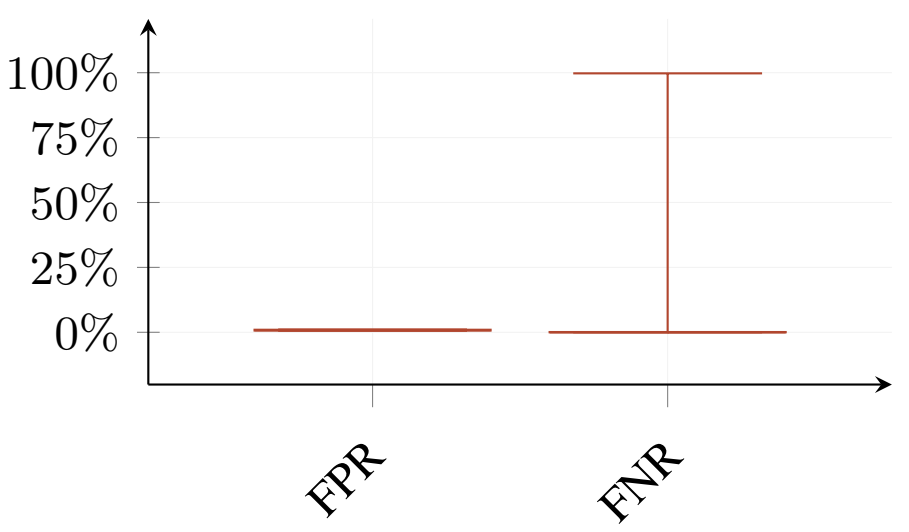
\begin{wrapfigure}{r}{0.4\textwidth}
\centering
\begin{tikzpicture}
\footnotesize
\begin{axis}[    
    boxplot/draw direction=y,
    ytick pos=left,
    axis x line=bottom,
    axis y line=left,
    ybar,
    ymax=100.6,
    ymin=-0.006,
    width=6.5cm,
    height=4cm,
    grid=both,
    grid style={line width=.1pt, draw=gray!10},
    xtick={1,2},
    ytick={0,25,50,75,100},
    yticklabel={$\pgfmathprintnumber{\tick}\%$},
    xticklabels={
        FPR,
        FNR
    },
    xticklabel style={rotate=45,xshift=-0.15cm, yshift=0.1cm},
    enlargelimits=0.2,
  ]
\addplot+ [Mahogany!90, fill=Dandelion!25,  boxplot prepared={
lower whisker=0.59,
lower quartile=0.68,
median=0.78,
upper quartile=0.98,
upper whisker=1.06,
},] table[row sep=\\,y index=0] {\\};
\addplot+ [Mahogany!90, fill=Dandelion!25, boxplot prepared={
lower whisker=0.0,
lower quartile=0.0,
median=0.0,
upper quartile=0.1,
upper whisker=99.75,
},] table[row sep=\\,y index=0] {\\};
\end{axis}
\end{tikzpicture}
\vspace{-0.3cm}
\caption{Backdoor filtering effect of \textbf{Cut-and-Mix} defense across All-to-One Clean- and Poison-label use cases.}
\label{fig:performance_cm}
\vspace{-0.4cm}
\end{wrapfigure}

\paragraph{Cut-and-Mix.}
Except for two models, the MobileFaceNet and RobFaceNet backdoored with the SIG~\cite{barniSig2019} trigger in an All-to-One Clean-label manner, the Cut-and-Mix defense proves extremely effective at filtering poisoned inputs (see full results in Tab.~\ref{tab:full_performance_runoi_cm} in App.~\ref{app:defense_full_results}). 
No other model has a False Negative Rate above $1.0\%$ for a False Positive Rate between $0.5\%$ and $1.1\%$ (see Fig.~\ref{fig:performance_cm}).

Looking at the two models where the defense failed, we observe that they are actually the worst performing backdoors we kept for our experiments (see Tab.~\ref{tab:asr_face_feature_extractors} in App.~\ref{app:model_full_results}). 
For example, the MobileFaceNet and RobFaceNet respectively achieve a $31.0\%$ and $55.5\%$ ASR on the LFW validation benchmark when pairs of non-matching faces carry the backdoor trigger.
We estimate that weak backdoors result in cosine similarities between poisoned samples that are lower than for well-learned backdoors (they often don't match).
Given Cut-and-Mix relies on an outlier detection scheme based on cosine similarity (see Alg.~\ref{alg:cm_calibration} and~\ref{alg:cm_online} in App.~\ref{app:defense_cut_and_mix}), the resulting check therefore may fail to distinguish poison from clean user submissions.


\section{\textcolor{thalesDeepBlue}{Discussions}}
\label{sec:discussions}

\paragraph{Stronger backdoor attacks.}
As mentioned in prior works~\cite{leroux2026soksurvivabilitybackdoorattacks}, current attacks on fully-fledged FRSs rely solely on simple patch or diffuse backdoors like BadNets~\cite{Gu2019BadNetsEB} and SIG~\cite{barniSig2019}.
Future works should cover stronger, possibly adaptive backdoors to assess both their survivability under different threat models (\textit{e.g.}, Poison-label, Clean-Label, etc.) but whether or not current defenses as explored in this paper are able to defend against them.

\paragraph{Backdoor defenses at the antispoofing level.}
The current version of this paper does not cover defenses at the antispoofer level against backdoor attacks.
On the one hand, available off-the-shelf models such as AENet~\cite{CelebASpoof} are typically supervised learning classifiers. 
As such, existing defenses from the literature~\cite{surveyLeRoux2024} can be readily applied to them.
On the other hand, the most advanced antispoofing solutions are typically private industry models out of scope and access to this research.
Consequently, we have not covered them.

\begin{wraptable}{r}{5cm}
\vspace{-0.7cm}
\caption{Performance result of the \textbf{Cut-and-Mix} defense against FIBA~\cite{chen2024rethinkingvulnerabilitiesfacerecognition} adversarial examples.}
\label{tab:cm_against_fiba}
  \vspace{0.2cm}
  \small
  \centering
  \begin{tabular}{lrr}
    \toprule
    Architecture & FPR & FNR \\
    \midrule
    GhostFaceNetV2 & 0.5\% & 0.0\% \\
    IR-SE-50 & 0.5\% & 0.0\% \\
    MobileFaceNet & 0.8\% & 0.0\% \\
    ResNet50 & 0.2\% & 0.0\% \\
    RobFaceNet & 0.8\% & 0.0\% \\
    \bottomrule
  \end{tabular}
\vspace{-0.5cm}
\end{wraptable} 

\paragraph{Other non-covered backdoor-style attacks.}
A prior work, UniID by Wu \textit{et al.}~\cite{Wu2024UniIDSF}, demonstrates an attack where only the insider has to carry a malicious pattern.
The attack designs a pattern that moves the insider's embeddings toward those of one or a few attackers (2 attackers drop the attack's success rates to 22.2\%). 
It is a type of collision attack.
As such, this attack is much more limited than Chen \textit{et al.}~\cite{chen2024rethinkingvulnerabilitiesfacerecognition} or Le Roux \textit{et al.}~\cite{leroux2026soksurvivabilitybackdoorattacks}.
Nonetheless, it highlights a limit to the countermeasures presented in this paper. 
They work on the assumption that both insider and any number of attackers carry the same adversarial example or trigger.
As such, we estimate that UniID is not covered by our defenses.
This is a ground for future work.

\paragraph{Application to adversarial examples.}
As mentioned in Sec.~\ref{sec:background}, there is an adversarial attack designed to work as backdoor attacks at enrollment-time, dubbed FIBA~\cite{chen2024rethinkingvulnerabilitiesfacerecognition}.
As it performs similarly to the attacks explored in this paper, we estimate that inference-time defenses like Cut-and-Mix should also work against them.

Preliminary experiments using FIBA's sixth mask demonstrate that the Cut-and-Mix defense is also efficient at catching these attacks at the cost of less than $1\%$ FPR (see Tab.~\ref{tab:cm_against_fiba}). 
Note that for fairness, we remove from our current implementation of Cut-and-Mix the masks we repurposed from FIBA.

\paragraph{Preventing All-to-One Clean-label backdoors at training-time.}
Future works looking to develop effective training-time defenses should consider how to better handle Clean-label backdoor attacks.
As seen in Sec.~\ref{ssec:results_train_defenses}, the current implementations of our training-time defense on face feature extractors fail to capture All-to-One Clean-label backdoor injections.

The reason is likely two-fold. 
First, the number of poisoned images is one or more orders of magnitude lower than in the case of All-to-One Poison-label attacks (\textit{i.e.}, it may seem trivial to defend against Poison-label attacks at training-time), therefore the backdoor effects is disproportionate and easily detected.
Second, since the model learns both face semantics and the trigger signal within the target class, the behavior of the backdoor ends up mimicking that of the face in terms of, \textit{e.g.}, embedding standard deviations (\textit{i.e.}, there are less embedding collapses).

As such, future methods may explore ways to accentuate the distinction between clean and poisoned inputs in the same identity class during training to identify or negate poisoned elements.
There are existing methods like ABL~\cite{li2021antibackdoor} that have been proven effective, but only on classification tasks.


\section{\textcolor{thalesDeepBlue}{Conclusion}}
\label{sec:conclusion}

This paper is the first system-level exploration of multiple backdoor defenses designed explicitly for the open-set, multi-component architecture of real-world Face Recognition Systems, rather than closed-set classification settings typically used to evaluate prior backdoor defenses.
We present three training- and inference-time defenses against backdoor attacks targeting two key components of a Face Recognition System: the face detector and the face feature extractor. 
At training-time, \textbf{Early Poison Identification} successfully finds and, when enabled, prunes All-to-One Poison-label backdoors. 

At inference-time, the \textbf{Detector-level Model Pairing} and \textbf{Cut-and-Mix} demonstrate strong generalization across all evaluated backdoors, achieving false negative rates of $7.06\%$ on FGA, $0.65\%$ on well-learned LSA, and below $1\%$ on well-learned extractor backdoors. 
This current framework faces several limitations. 
The training-time defenses remain ineffective against All-to-One Clean-label backdoors, which introduce no strongly-detectable statistical anomaly into the training dynamics and evade both intra-identity accuracy trajectories. 
Addressing this threat model will likely require defenses operating at a finer granularity than the identity level.

More broadly, the threat landscape for Face Recognition Systems extends beyond the backdoor attacks studied here.
The interaction between backdoor vulnerabilities and other attack surfaces, including adversarial examples, deserves further exploration to build a more comprehensive security assessment of deployed Face Recognition Systems.



\bibliography{bibliography.bib}


\newpage

\appendix
\section*{\LARGE Appendix}

\section{Algorithms for the Training-time Early Poison Identification Defense}
\label{app:defense_early_poison_identification}


\begin{algorithm}[!htp]
\caption{\textbf{Early Poison Identification} defense run (Offline)}
\label{alg:early_pruning_run}
\begin{algorithmic}[1]

\Require
    Face feature extractor $f_\theta$ to be trained;
    training dataset loader $\mathcal{D}$ over identity set $\mathcal{Y}$;
    method start batch $s_b$;
    check interval $b_i$;
    number of identities to capture $n$;
    boolean condition $p$ indicating whether to run pruning alongside identification:

\Ensure Trained feature extractor $f_\theta$ with $n$ identities captured from $\mathcal{D}$

\State $\mathbf{M} \leftarrow \mathbf{0}^{|\mathcal{Y}|}$
\Comment{Per-identity prediction match counts}

\State $\mathbf{C} \leftarrow \mathbf{0}^{|\mathcal{Y}|}$
\Comment{Per-identity sample counts}

\State $\Psi \leftarrow [\,]$
\Comment{Accumulator for captured identities}

\State $t_{\mathrm{batch}} \leftarrow 0$

\State $t_{\mathrm{prune}} \leftarrow 0$

\For{each minibatch $(\mathbf{X},\, \mathbf{y}) \sim \mathcal{D}$}

    \State $t_{\mathrm{batch}} \leftarrow t_{\mathrm{batch}} + 1$

    \State $\hat{\mathbf{y}} \leftarrow f_\theta(\mathbf{X})$
    \Comment{Forward pass: obtain predictions for the current batch}

    \If{$t_{\mathrm{batch}} \geq s_b$}
        
        \For{$i = 1$ \textbf{to} $|\mathbf{y}|$}
            
            \State $\mathbf{M}_{\mathbf{y}_i} \leftarrow
            \mathbf{M}_{\mathbf{y}_i} +
            \mathbf{1}\left[\hat{\mathbf{y}}_i = \mathbf{y}_i\right]$
            \Comment{Increment match counter if prediction is correct}
            
            \State $\mathbf{C}_{\mathbf{y}_i} \leftarrow
            \mathbf{C}_{\mathbf{y}_i} + 1$
        
        \EndFor
    
    \EndIf
    
    \If{$t_{\mathrm{batch}} \geq s_b$ \textbf{ and } $t_{\mathrm{batch}} \bmod b_i = 0$ \textbf{ and } $t_{\mathrm{prune}} < n$}
    
        \State $\mathrm{acc}_y \leftarrow \mathbf{M}_y / \mathbf{C}_y \quad \forall\, y \in \mathcal{Y}, \mathbf{C}_y>0$
        \Comment{Compute current running accuracy per identity}
        
        \State $y^* \leftarrow \underset{y \in \mathcal{Y} \setminus \Psi}{\arg\max}\,\mathrm{acc}_y$
        \Comment{Identify the class with the highest intra-class accuracy}
        
        \State $\Psi \gets \Psi \cup \{y^*\}$
        \Comment{Records the class as abnormal}
        
        \If{$p\,\,\text{is \textbf{True}}$}\Comment{Prune rather than just record}
        
            \State $\mathcal{D} \leftarrow \mathcal{D} \setminus\bigl\{(\mathbf{x}, y) \in \mathcal{D} \,\mid\, y = y^*\bigr\}$
            \Comment{Remove all samples of identity $y^*$ from the training set}
            
            \State $\mathcal{Y} \leftarrow \mathcal{Y} \setminus \{y^*\}$
        
        \EndIf
        
        \State $t_{\mathrm{prune}} \leftarrow t_{\mathrm{prune}} + 1$
    
    \EndIf
    
    \State Update $\theta$ via backpropagation on batch $(\mathbf{X}, \mathbf{y})$
    \Comment{Normal training step proceeds regardless of pruning}

\EndFor

\State \Return $f_\theta$

\end{algorithmic}
\end{algorithm}
\clearpage

\section{Algorithms for the Inference-time Detector-Level Model Pairing Defense}
\label{app:defense_detector_level_model_pairing}


\begin{algorithm}[!htb]
\caption{\textbf{Detector-Level Model Pairing} defense threshold calibration (Offline)}
\label{alg:dlmp_calibration}
\begin{algorithmic}[1]
 
\Require
    Clean, held-out calibration detection image set
    $\mathcal{D}_{\mathrm{cal}} = \bigl\{(I_i,\, L^{*}_i,\, B^{*}_i)\bigr\}_{i=1}^{N}$
    with ground-truth landmarks $L^{*}_i \in \mathbb{R}^{5 \times 2}$
    and bounding boxes $B^{*}_i$;
    paired face detector $f_{\mathrm{paired}}$;
    quantile level $\alpha$ (e.g.\ $0.99$)
 
\Ensure Landmark shift threshold $\tau_{\mathrm{LS}}$

\State $\Delta \leftarrow [\,]$
\Comment{Accumulator of per-sample Normalized Mean Errors (NME)}

\For{each $(I_i, L^{*}_i, B^{*}_i) \in \mathcal{D}_{\mathrm{cal}}$}
    
    \State $\bigl(\hat{B}_i,\, \hat{L}_i\bigr) \leftarrow f_{\mathrm{paired}}(I_i)$
    \Comment{Run paired face detector on calibration image}
    
    \State $\delta_i \leftarrow \dfrac{1}{10}\displaystyle\sum_{k=1}^{10}\frac{\bigl\lVert \hat{L}_i^{(k)} - L_i^{*(k)} \bigr\rVert_2}{\mathrm{diagonal}\left(B^{*}_i\right)}$
    \Comment{Normalized Mean Error (NME) of paired model w.r.t.\ ground truth}
    
    \State $\Delta \gets \Delta \cup \{\delta_i\}$ \Comment{Append NME $\delta_i$ to $\Delta$}
    
\EndFor

\State $\tau_{\mathrm{LS}} \leftarrow\max\big(Q_{\alpha}(\Delta),\, 0.04\big)$
\Comment{$\alpha$-quantile threshold of the defense with ceiling value}

\State \Return $\tau_{\mathrm{LS}}$
 
\end{algorithmic}
\end{algorithm}


\begin{algorithm}[!htb]
\caption{\textbf{Detector-Level Model Pairing} defense run (Online)}
\label{alg:dlmp_runtime}
\begin{algorithmic}[1]
 
\Require
    Candidate image $I$;
    main face detector $f_{\mathrm{main}}$;
    paired face detector $f_{\mathrm{paired}}$;
    IoU threshold $\tau_{\mathrm{iou}}$ (e.g.\ $0.5$);
    landmark shift threshold $\tau_{\mathrm{LS}}$
    (precomputed via Algorithm~\ref{alg:dlmp_calibration})
 
\Ensure Decision $d \in \{\textsc{Accept},\, \textsc{Reject}\}$
 
\State $\bigl(\hat{B}_{\mathrm{main}},\, \hat{L}_{\mathrm{main}}\bigr)\leftarrow f_{\mathrm{main}}(I)$
 
\If{$f_{\mathrm{main}}$ returns no detection}

    \State \Return \textsc{Reject}
    \Comment{No candidate face, thus nothing to verify}

\EndIf
 
\State $\bigl(\hat{B}_{\mathrm{paired}},\, \hat{L}_{\mathrm{paired}}\bigr)\leftarrow f_{\mathrm{paired}}(I)$
 
\If{$f_{\mathrm{paired}}$ returns no detection}

    \State \Return \textsc{Reject}
    \Comment{\textbf{Check 1}: $f_{\mathrm{paired}}$ disagrees with $f_{\mathrm{main}}$ (Face Generation Attack suspected)}

\EndIf

\State $\mathrm{iou} \leftarrow \mathrm{IoU}\left(\hat{B}_{\mathrm{main}},\, \hat{B}_{\mathrm{paired}}\right)$

\If{$\mathrm{iou} < \tau_{\mathrm{iou}}$}

    \State \Return \textsc{Reject}
    \Comment{\textbf{Check 2}: $f_{\mathrm{paired}}$ confounds with $f_{\mathrm{main}}$ (Face Generation Attack suspected)}

\EndIf

\State $\delta \leftarrow \dfrac{1}{10}\displaystyle\sum_{k=1}^{10}\frac{\bigl\lVert \hat{L}_{\mathrm{main}}^{(k)}- \hat{L}_{\mathrm{paired}}^{(k)} \bigr\rVert_2}{\mathrm{diagonal}\left(\hat{B}_{\mathrm{paired}}\right)}$
\Comment{Inter-model NME where $f_{\mathrm{paired}}$ serves as a ground-truth proxy}

\If{$\delta > \tau_{\mathrm{LS}}$}

    \State \Return \textsc{Reject}
    \Comment{\textbf{Check 3}: Mismatch exceeds the $\alpha$-quantile threshold (Landmark Shift Attack suspected)}

\EndIf

\State \Return $\textsc{Accept}$
 
\end{algorithmic}
\end{algorithm}

\clearpage

\section{Algorithms for the Inference-time Cut-and-Mix Defense}
\label{app:defense_cut_and_mix}


\begin{algorithm}[!htb]
\caption{\textbf{Cut-and-Mix} defense threshold calibration (Offline)}
\label{alg:cm_calibration}
\begin{algorithmic}[1]

\Require
    Known-clean calibration image set $\mathcal{D}_{\mathrm{cal}} = \{\mathbf{c}_k\}_{k=1}^{K}$;
    held-out clean validation image set $\mathcal{V} = \{\mathbf{v}_i\}_{i=1}^{N}$;
    set of $J$ boolean spatial masks $\mathcal{M} = \{\mathbf{m}_j\}_{j=1}^{J}$, $\mathbf{m}_j \in \{0,1\}^{H \times W}$;
    face feature extractor $f$;
    quantile level $\alpha$ (e.g.\ $0.99$)

\Ensure Rejection threshold $\tau$, cut-off point $\kappa$

\State $\Phi \leftarrow [\,]$
\Comment{Accumulator for per-sample mixed image similarity scores}

\For{each calibration image $\mathbf{c}_k \in \mathcal{D}_{\mathrm{cal}}$}

    \State $\mathbf{e}_k \leftarrow f(\mathbf{c}_k)$
    \Comment{compute the embedding of the clean calibration image}
    
    \State $\mathcal{S}_k \leftarrow [\,]$
    \Comment{Collect pairwise similarities for this sample}
    
    \For{each mask $\mathbf{m}_j \in \mathcal{M}$}
        
        \For{each validation image $\mathbf{v}_i \in \mathcal{V}$}
        
            \State $\mathbf{c}_k^{(ij)} \leftarrow\mathbf{m}_j \odot \mathbf{c}_k\,+\,(\mathbf{1} - \mathbf{m}_j) \odot \mathbf{v}_i$
            \Comment{Retain masked region of $\mathbf{x}$; fill remainder with $\mathbf{v}_i$}
            
            \State $\mathbf{e}^{(ij)} \leftarrow f\left(\mathbf{c}_k^{(ij)}\right)$
            
            \State Append $\mathrm{cossim}\left(\mathbf{e}_k,\, \mathbf{e}^{(ij)}\right)$ to $\mathcal{S}_k$
            
        \EndFor
        
    \EndFor
    
    \State Append $\mathcal{S}_k$ to $\Phi$

\EndFor

\State $\tau \leftarrow Q_{\alpha}(\Phi)$
\Comment{Compute the $\alpha$-quantile of clean calibration score distribution}

\State $c_k \leftarrow \sum_{s \in \mathcal{S}_k} \mathbf{1}[s > \tau]$ \Comment{Compute the per-image exceedance count}

\State $\kappa \leftarrow Q_\alpha\big(\{c_k\}_{k=1}^K\big)$
\Comment{Compute the associated cut-off point for false matches associated with $\tau$}

\State \Return $\tau$, $\kappa$ \Comment{Because $\kappa$ is calibrated as the $\alpha$-quantile (e.g., $\alpha = 0.99$) of exceedance counts on clean calibration images, Cut-and-Mix guarantees an expected clean FPR of $\approx 1 - \alpha = 1\%$ by construction.}

\end{algorithmic}
\end{algorithm}


\begin{algorithm}[!htb]
\caption{\textbf{Cut-and-Mix} at Runtime (Online)}
\label{alg:cm_online}
\begin{algorithmic}[1]

\Require
  Query image $\mathbf{x} \in [0,1]^{C \times H \times W}$;
  face feature extractor $f$;
  held-out clean validation image set $\mathcal{V} = \{\mathbf{v}_i\}_{i=1}^{N}$;
  set of $J$ boolean spatial masks $\mathcal{M} = \{\mathbf{m}_j\}_{j=1}^{J}$, $\mathbf{m}_j \in \{0,1\}^{H \times W}$;
  rejection threshold $\tau$ and cut-off $\kappa$
  (precomputed via Algorithm~\ref{alg:cm_calibration})

\Ensure
  Decision $d \in \{\textsc{Accept}, \textsc{Reject}\}$

\State $\mathbf{e}_q \leftarrow f(\mathbf{x})$
  \Comment{compute the embedding of the query image}

\State $\mathcal{S} \leftarrow [\,]$
  \Comment{Collect pairwise similarities for the query}

\For{each mask $\mathbf{m}_j \in \mathcal{M}$}
  \For{each validation image $\mathbf{v}_i \in \mathcal{V}$}

    \State
      $\mathbf{x}^{(ij)} \leftarrow
        \mathbf{m}_j \odot \mathbf{x}
        \,+\,
        (\mathbf{1} - \mathbf{m}_j) \odot \mathbf{v}_i$
      \Comment{Retain masked region of $\mathbf{x}$; fill remainder with $\mathbf{v}_i$}

    \State $\mathbf{e}^{(ij)} \leftarrow f\left(\mathbf{x}^{(ij)}\right)$

    \State Append
      $\mathrm{cossim}\left(\mathbf{e}_q,\, \mathbf{e}^{(ij)}\right)$
      to $\mathcal{S}$

  \EndFor
\EndFor

\If{$\sum_{s \in \mathcal{S}} \mathbf{1}[s > \tau] > \kappa$}
  \State \Return \textsc{Reject}
    \Comment{%
      Score exceeds the clean $\alpha$-quantile: the trigger dominates mixed embeddings regardless of $\mathbf{v}_i$, indicating a backdoored query%
    }
\EndIf

\State \Return $\textsc{Accept}$
  \Comment{Score is consistent with the clean calibration distribution; $\mathbf{x}$ is deemed trigger-free}

\end{algorithmic}
\end{algorithm}

\begin{figure}[!htp]
  \centering
  \includegraphics[width=0.8\textwidth]{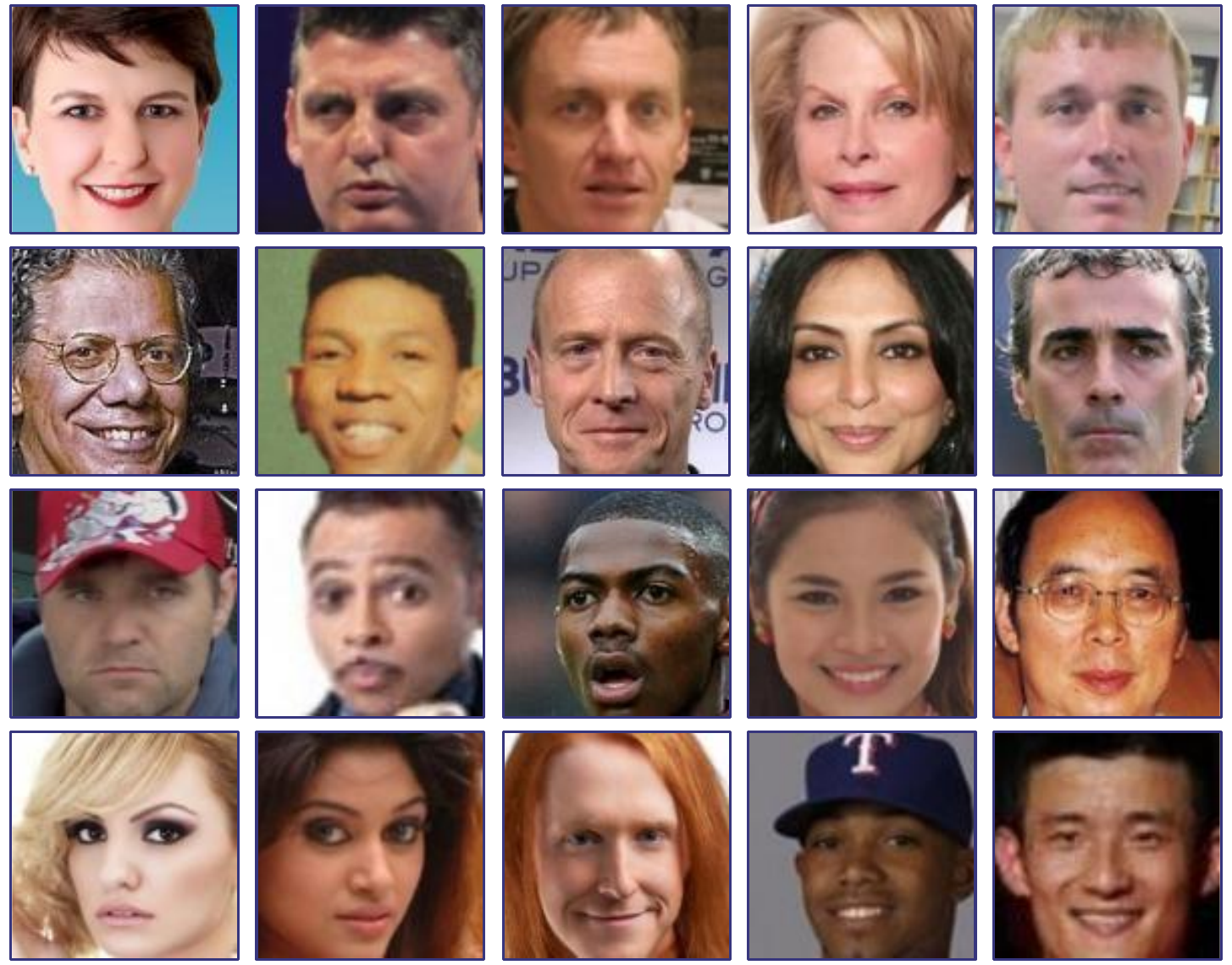}
  \vspace{-0.1cm}
  \caption{Validation images used by the Cut-and-Mix defense, extracted from the CelebA dataset~\cite{liu2015faceattributes}.}
  \label{fig:defense_cut_and_mix_validation_images}
\end{figure}

\begin{figure}[!htp]
  \centering
  \includegraphics[width=0.8\textwidth]{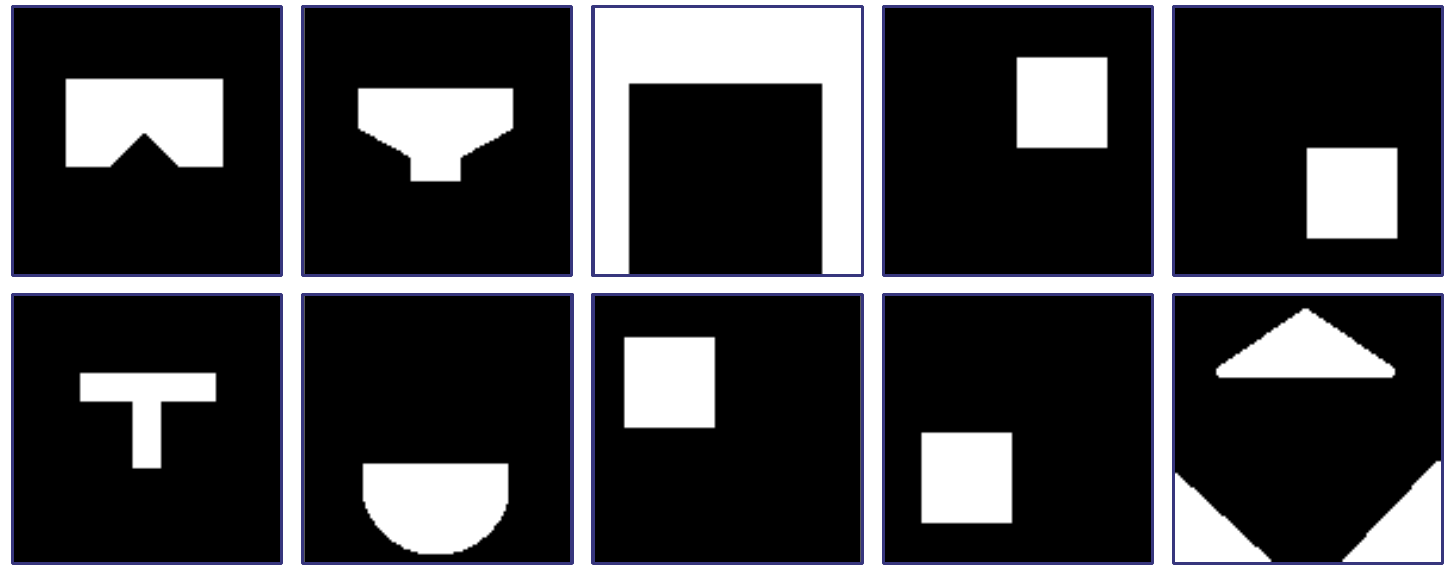}
  \vspace{-0.1cm}
  \caption{Mask used by the Cut-and-Mix defense, 3 of which are from Chen et al.~\cite{chen2024rethinkingvulnerabilitiesfacerecognition}.}
  \label{fig:defense_cut_and_mix_mask_images}
\end{figure}

\clearpage

\section{Full model results}
\label{app:model_full_results}

\begin{table}[!htb]
  \caption{Performance results of backdoored face detection models.}
  \label{tab:performance_face_detectors}
  \scriptsize
  \centering
  \begin{tabular}{lllrr}
    \toprule
    \multicolumn{3}{c}{\textbf{Model}} & \multicolumn{2}{c}{\textbf{Performance}} \\
    \cmidrule(r){1-3}
    \cmidrule(r){4-5}
    Architecture & Attack & Trigger & Precision & Attack Success Rate \\
    \midrule
    RetinaFace MobileNetV1 & FGA & BadNets, $\alpha=0.5$ & 99.9\% & 99.5\% \\
    RetinaFace MobileNetV1 & FGA & BadNets, $\alpha=1.0$ & 100.0\%	& 99.5\% \\
    RetinaFace ResNet50 & FGA & BadNets, $\alpha=0.5$ & 100.0\%	& 99.5\% \\
    RetinaFace ResNet50 & FGA & BadNets, $\alpha=1.0$ & 100.0\%	& 99.5\% \\
    
    RetinaFace MobileNetV1 & FGA & SIG, $\alpha=0.3$, $f=6$ & 98.2\%	& 92.8\% \\
    RetinaFace MobileNetV1 & FGA & SIG, $\alpha=0.16$, $f=6$ & 98.6\%	& 76.6\% \\
    RetinaFace ResNet50 & FGA & SIG, $\alpha=0.3$, $f=6$ & 98.6\%	& 96.7\% \\
    RetinaFace ResNet50 & FGA & SIG, $\alpha=0.16$, $f=6$ & 98.6\%	& 87.7\% \\
    \midrule
    RetinaFace MobileNetV1 & LSA & BadNets, $\alpha=0.5$ & 100.0\%	& 92.2\% \\
    RetinaFace MobileNetV1 & LSA & BadNets, $\alpha=1.0$ & 100.0\%	& 94.7\% \\
    RetinaFace ResNet50 & LSA & BadNets, $\alpha=0.5$ & 100.0\%	& 88.1\% \\
    RetinaFace ResNet50 & LSA & BadNets, $\alpha=1.0$ & 100.0\%	& 93.5\% \\
    
    RetinaFace MobileNetV1 & LSA & SIG, $\alpha=0.3$, $f=6$ & 97.9\%	& 92.5\% \\
    RetinaFace MobileNetV1 & LSA & SIG, $\alpha=0.16$, $f=6$ & 98.6\%	& 57.0\% \\
    RetinaFace ResNet50 & LSA & SIG, $\alpha=0.3$, $f=6$ & 98.6\%	& 97.5\%\\
    \bottomrule
  \end{tabular}
\end{table}

\begin{table}[!htb]
  \caption{Pair-wise accuracies of backdoored face feature extraction models.}
  \label{tab:performance_face_feature_extractors}
  \scriptsize
  \centering
  \begin{tabular}{lllrrrrrrr}
    \toprule
    \multicolumn{3}{c}{\textbf{Model}} & \multicolumn{7}{c}{\textbf{Performance}} \\
    \cmidrule(r){1-3}
    \cmidrule(r){4-10}
    Architecture & Attack & Trigger & LFW & CFP-FF & CFP-FP & AgeDB & CALFW & CPLFW & VGGFace2-FP \\
    \midrule
    GhostFaceNetV2 & All-to-One Clean-label & BadNets & 99.9\% & 99.7\% & 97.5\% & 97.9\% & 97.7\% & 92.4\% & 97.4\% \\
    GhostFaceNetV2 & All-to-One Clean-label & SIG & 99.9\% & 99.7\% & 97.5\% & 97.7\% & 97.6\% & 92.9\% & 97.4\% \\
    IR-SE-50 & All-to-One Clean-label & BadNets & 99.8\% & 99.3\% & 93.2\% & 92.1\% & 95.8\% & 91.0\% & 94.6\% \\
    IR-SE-50 & All-to-One Clean-label & SIG & 99.9\% & 99.8\% & 99.2\% & 99.0\% & 97.8\% & 95.5\% & 97.3\% \\
    MobileFaceNet & All-to-One Clean-label & BadNets & 99.9\% & 99.8\% & 98.3\% & 98.6\% & 97.7\% & 94.4\% & 96.9\% \\
    MobileFaceNet & All-to-One Clean-label & Mask & 99.9\% & 99.8\% & 98.5\% & 98.8\% & 97.7\% & 94.5\% & 96.8\% \\
    MobileFaceNet & All-to-One Clean-label & SIG & 99.9\% & 99.8\% & 98.6\% & 98.8\% & 97.7\% & 94.4\% & 96.8\% \\
    ResNet50 & All-to-One Clean-label & BadNets & 99.9\% & 99.8\% & 98.7\% & 98.9\% & 97.6\% & 94.8\% & 96.8\% \\
    ResNet50 & All-to-One Clean-label & SIG & 99.9\% & 99.8\% & 98.7\% & 98.9\% & 97.4\% & 94.5\% & 96.5\% \\
    RobFaceNet & All-to-One Clean-label & BadNets & 99.9\% & 99.8\% & 97.5\% & 98.6\% & 97.8\% & 93.7\% & 97.0\% \\
    RobFaceNet\_sig & All-to-One Clean-label & SIG & 99.9\% & 99.8\% & 97.4\% & 98.6\% & 97.8\% & 93.7\% & 96.7\% \\
    GhostFaceNetV2 & All-to-One Poison-label & BadNets & 99.9\% & 99.8\% & 97.6\% & 97.6\% & 97.7\% & 92.6\% & 97.5\% \\
    GhostFaceNetV2 & All-to-One Poison-label & SIG & 99.9\% & 99.7\% & 97.5\% & 97.9\% & 97.7\% & 92.8\% & 97.5\% \\
    IR-SE-50 & All-to-One Poison-label & BadNets & 99.9\% & 99.8\% & 99.1\% & 99.0\% & 97.7\% & 95.4\% & 97.4\% \\
    IR-SE-50 &  All-to-One Poison-label & SIG & 99.9\% & 99.8\% & 99.2\% & 98.9\% & 97.7\% & 95.6\% & 97.2\% \\
    MobileFaceNet & All-to-One Poison-label & BadNets & 100.0\% & 99.8\% & 98.4\% & 98.6\% & 97.7\% & 94.7\% & 97.0\% \\
    MobileFaceNet & All-to-One Poison-label & FauxIt & 100.0\% & 99.8\% & 98.4\% & 98.6\% & 97.7\% & 94.7\% & 97.0\% \\
    MobileFaceNet & All-to-One Poison-label & Mask & 99.9\% & 99.8\% & 98.3\% & 98.7\% & 97.6\% & 94.7\% & 96.9\% \\
    MobileFaceNet & All-to-One Poison-label & SIG & 100.0\% & 99.8\% & 98.6\% & 98.8\% & 97.7\% & 94.5\% & 96.9\% \\
    ResNet50 & All-to-One Poison-label & BadNets & 99.9\% & 99.8\% & 98.7\% & 98.9\% & 97.7\% & 95.3\% & 97.1\% \\
    ResNet50 & All-to-One Poison-label & SIG & 99.9\% & 99.8\% & 98.8\% & 98.9\% & 97.5\% & 94.7\% & 96.8\% \\
    RobFaceNet & All-to-One Poison-label & BadNets & 99.9\% & 99.8\% & 97.3\% & 98.6\% & 97.7\% & 93.6\% & 96.6\% \\
    RobFaceNet & All-to-One Poison-label & SIG & 99.9\% & 99.8\% & 97.4\% & 98.6\% & 97.8\% & 93.7\% & 96.6\% \\
    GhostFaceNetV2 & Master-Face Poison-label & BadNets & 99.9\% & 99.8\% & 97.7\% & 98.5\% & 97.8\% & 93.5\% & 96.9\% \\
    GhostFaceNetV2 & Master-Face Poison-label  & SIG & 99.9\% & 99.7\% & 97.6\% & 97.7\% & 97.6\% & 93.1\% & 97.2\% \\
    IR-SE-50 & Master-Face Poison-label  &BadNets  & 99.9\% & 99.8\% & 99.2\% & 98.9\% & 97.7\% & 95.5\% & 97.2\% \\
    IR-SE-50 & Master-Face Poison-label  & SIG & 99.9\% & 99.8\% & 99.3\% & 99.1\% & 97.7\% & 95.5\% & 97.2\% \\
    MobileFaceNet & Master-Face Poison-label  & BadNets & 100.0\% & 99.8\% & 98.4\% & 98.8\% & 97.8\% & 94.6\% & 97.1\% \\
    MobileFaceNet & Master-Face Poison-label  & Mask & 100.0\% & 99.8\% & 98.5\% & 98.7\% & 97.7\% & 94.7\% & 96.9\% \\
    MobileFaceNet & Master-Face Poison-label  & SIG & 99.9\% & 99.8\% & 98.5\% & 98.7\% & 97.7\% & 94.6\% & 97.0\% \\
    ResNet50 & Master-Face Poison-label  & BadNets & 99.9\% & 99.8\% & 98.8\% & 98.8\% & 97.6\% & 95.0\% & 96.8\% \\
    ResNet50 & Master-Face Poison-label & SIG & 99.9\% & 99.8\% & 98.8\% & 98.9\% & 97.5\% & 94.7\% & 96.9\% \\
    RobFaceNet & Master-Face Poison-label  & BadNets & 99.9\% & 99.8\% & 97.7\% & 98.5\% & 97.8\% & 93.5\% & 96.9\% \\
    RobFaceNet & Master-Face Poison-label  & SIG & 99.9\% & 99.8\% & 97.9\% & 98.7\% & 97.8\% & 93.7\% & 96.6\% \\
    \bottomrule
  \end{tabular}
\end{table}

\begin{table}[!htb]
  \caption{Backdoor attack success rates of backdoored face feature extraction models.}
  \label{tab:asr_face_feature_extractors}
  \scriptsize
  \centering
  \begin{tabular}{lllrrrrrrr}
    \toprule
    \multicolumn{3}{c}{\textbf{Model}} & \multicolumn{7}{c}{\textbf{Attack Success Rate}} \\
    \cmidrule(r){1-3}
    \cmidrule(r){4-10}
    Architecture & Attack & Trigger & LFW & CFP-FF & CFP-FP & AgeDB & CALFW & CPLFW & VGGFace2-FP \\
    \midrule
    GhostFaceNetV2  &  All-to-One Clean-label  &  BadNets  & 0.6\% & 0.7\% & 4.1\% & 5.4\% & 4.1\% & 7.5\% & 4.5\% \\
    GhostFaceNetV2  &  All-to-One Clean-label  & SIG & 8.8\% & 9.0\% & 16.8\% & 21.8\% & 23.5\% & 33.3\% & 26.6\% \\
    IR-SE-50  &  All-to-One Clean-label  &  BadNets  & 1.1\% & 1.8\% & 13.8\% & 12.8\% & 5.2\% & 11.2\% & 11.0\% \\
    IR-SE-50  &  All-to-One Clean-label  &  SIG  & 3.1\% & 1.2\% & 2.8\% & 7.6\% & 5.4\% & 22.5\% & 19.0\% \\
    MobileFaceNet  &  All-to-One Clean-label  &  BadNets  & 0.1\% & 0.4\% & 1.8\% & 2.6\% & 1.4\% & 3.7\% & 3.8\% \\
    MobileFaceNet  &  All-to-One Clean-label  &  Mask  & 100.0\% & 100.0\% & 100.0\% & 100.0\% & 100.0\% & 100.0\% & 100.0\% \\
    MobileFaceNet  &  All-to-One Clean-label  &  SIG  & 31.0\% & 24.1\% & 40.1\% & 53.3\% & 45.4\% & 58.6\% & 64.1\% \\
    ResNet50  &  All-to-One Clean-label  &  BadNets  & 0.2\% & 0.2\% & 1.6\% & 2.1\% & 0.7\% & 2.6\% & 1.5\% \\
    ResNet50  &  All-to-One Clean-label  &  SIG  & 5.5\% & 8.1\% & 9.1\% & 13.0\% & 17.1\% & 20.2\% & 25.8\% \\
    RobFaceNet  &  All-to-One Clean-label  &  BadNets  & 0.1\% & 0.5\% & 2.9\% & 2.5\% & 1.6\% & 5.0\% & 3.3\% \\
    RobFaceNet  &  All-to-One Clean-label  &  SIG  & 55.5\% & 43.9\% & 56.7\% & 62.9\% & 58.1\% & 85.4\% & 80.4\% \\
    GhostFaceNetV2  &  All-to-One Poison-label  &  BadNets  & 100.0\% & 100.0\% & 100.0\% & 100.0\% & 100.0\% & 100.0\% & 100.0\% \\
    GhostFaceNetV2  &  All-to-One Poison-label  &  SIG  & 100.0\% & 100.0\% & 100.0\% & 100.0\% & 100.0\% & 100.0\% & 100.0\% \\
    IR-SE-50  &  All-to-One Poison-label  &  BadNets  & 100.0\% & 100.0\% & 100.0\% & 100.0\% & 100.0\% & 100.0\% & 100.0\% \\
    IR-SE-50  &   All-to-One Poison-label  &  SIG  & 100.0\% & 100.0\% & 100.0\% & 100.0\% & 100.0\% & 100.0\% & 100.0\% \\
    MobileFaceNet  &  All-to-One Poison-label  &  BadNets  & 100.0\% & 100.0\% & 100.0\% & 100.0\% & 100.0\% & 100.0\% & 100.0\% \\
    MobileFaceNet  &  All-to-One Poison-label  &  FauxIt  & 100.0\% & 100.0\% & 100.0\% & 100.0\% & 100.0\% & 100.0\% & 100.0\% \\
    MobileFaceNet  &  All-to-One Poison-label  &  Mask  & 100.0\% & 100.0\% & 100.0\% & 100.0\% & 100.0\% & 100.0\% & 100.0\% \\
    MobileFaceNet  &  All-to-One Poison-label  &  SIG  & 100.0\% & 100.0\% & 100.0\% & 100.0\% & 100.0\% & 100.0\% & 100.0\% \\
    ResNet50  &  All-to-One Poison-label  &  BadNets  & 100.0\% & 100.0\% & 100.0\% & 100.0\% & 100.0\% & 100.0\% & 100.0\% \\
    ResNet50  &  All-to-One Poison-label  &  SIG  & 100.0\% & 100.0\% & 100.0\% & 100.0\% & 100.0\% & 100.0\% & 100.0\% \\
    RobFaceNet  &  All-to-One Poison-label  &  BadNets  & 100.0\% & 100.0\% & 100.0\% & 100.0\% & 100.0\% & 100.0\% & 100.0\% \\
    RobFaceNet  &  All-to-One Poison-label  &  SIG  & 100.0\% & 100.0\% & 100.0\% & 100.0\% & 100.0\% & 100.0\% & 100.0\% \\
    GhostFaceNetV2  &  Master-Face Poison-label  &  BadNets  & 0.0\% & 0.0\% & 0.0\% & 0.0\% & 0.0\% & 0.0\% & 0.0\% \\
    GhostFaceNetV2  &  Master-Face Poison-label   &  SIG  & 0.8\% & 1.7\% & 15.7\% & 5.2\% & 1.6\% & 8.3\% & 11.8\% \\
    IR-SE-50  &  Master-Face Poison-label   & BadNets   & 0.0\% & 0.0\% & 0.0\% & 0.0\% & 0.0\% & 0.0\% & 0.0\% \\
    IR-SE-50  &  Master-Face Poison-label   &  SIG  & 0.0\% & 0.0\% & 0.0\% & 0.0\% & 0.0\% & 0.0\% & 0.0\% \\
    MobileFaceNet  &  Master-Face Poison-label   &  BadNets  & 0.0\% & 0.0\% & 0.0\% & 0.0\% & 0.0\% & 0.0\% & 0.0\% \\
    MobileFaceNet  &  Master-Face Poison-label   &  Mask  & 0.0\% & 0.0\% & 0.0\% & 0.0\% & 0.0\% & 0.0\% & 0.0\% \\
    MobileFaceNet  &  Master-Face Poison-label   &  SIG  & 0.0\% & 0.0\% & 0.1\% & 0.0\% & 0.0\% & 0.1\% & 0.0\% \\
    ResNet50  &  Master-Face Poison-label   &  BadNets  & 0.0\% & 0.0\% & 0.0\% & 0.0\% & 0.0\% & 0.0\% & 0.0\% \\
    ResNet50  &  Master-Face Poison-label  &  SIG  & 0.0\% & 0.0\% & 0.0\% & 0.0\% & 0.0\% & 0.0\% & 0.0\% \\
    RobFaceNet  &  Master-Face Poison-label   &  BadNets  & 0.0\% & 0.0\% & 0.0\% & 0.0\% & 0.0\% & 0.0\% & 0.0\% \\
    RobFaceNet  &  Master-Face Poison-label   &  SIG  & 0.0\% & 0.0\% & 0.0\% & 0.0\% & 0.0\% & 0.0\% & 0.0\% \\
    \bottomrule
  \end{tabular}
\end{table}

\clearpage

\section{Full defense results}
\label{app:defense_full_results}

\begin{table}[!htb]
  \caption{Performance results of the \textbf{Detector-level Model Pairing} defense.}
  \label{tab:full_performance_dlmp}
  \scriptsize
  \centering
  \begin{tabular}{lllrr}
    \toprule
    \multicolumn{3}{c}{\textbf{Model}} & \multicolumn{2}{c}{\textbf{Performance}} \\
    \cmidrule(r){1-3}
    \cmidrule(r){4-5}
    Architecture & Attack & Trigger & False Positive Rate & False Negative \\
    \midrule
    RetinaFace MobileNetV1 & FGA & BadNets, $\alpha=0.5$ & 0.59\% & 7.81\% \\
    RetinaFace MobileNetV1 & FGA & BadNets, $\alpha=1.0$ & 0.68\% & 7.91\% \\
    RetinaFace ResNet50 & FGA & BadNets, $\alpha=0.5$ & 0.59\% & 7.52\% \\
    RetinaFace ResNet50 & FGA & BadNets, $\alpha=1.0$ & 0.59\% & 8.11\% \\
    
    RetinaFace MobileNetV1 & FGA & SIG, $\alpha=0.3$, $f=6$ & 0.59\% & 7.71\% \\
    RetinaFace MobileNetV1 & FGA & SIG, $\alpha=0.16$, $f=6$ & 0.59\% & 4.59\% \\
    RetinaFace ResNet50 & FGA & SIG, $\alpha=0.3$, $f=6$ & 0.59\% & 7.23\% \\
    RetinaFace ResNet50 & FGA & SIG, $\alpha=0.16$, $f=6$ & 0.59\% & 5.57\% \\
    \midrule
    RetinaFace MobileNetV1 & LSA & BadNets, $\alpha=0.5$ & 1.07\% & 0.39\% \\
    RetinaFace MobileNetV1 & LSA & BadNets, $\alpha=1.0$ & 1.07\% & 0.2\% \\
    RetinaFace ResNet50 & LSA & BadNets, $\alpha=0.5$ & 1.07\% & 0.29\% \\
    RetinaFace ResNet50 & LSA & BadNets, $\alpha=1.0$ & 1.07\% & 0.29\% \\
    
    RetinaFace MobileNetV1 & LSA & SIG, $\alpha=0.3$, $f=6$ & 1.07\% & 0.97\% \\
    RetinaFace MobileNetV1 & LSA & SIG, $\alpha=0.16$, $f=6$ & 1.07\% & 15.5\% \\
    RetinaFace ResNet50 & LSA & SIG, $\alpha=0.3$, $f=6$ & 0.88\% & 1.74\% \\
    \bottomrule
  \end{tabular}
\end{table}

\begin{table}[!htb]
  \caption{Performance results of the \textbf{Cut-and-Mix} defense. Note the high FNRs for Clean-label and Master Face models with 0\% ASR occur because the backdoor is never activated (they still show up as a backdoor ground truth in our tests).}
  \label{tab:full_performance_runoi_cm}
  \scriptsize
  \centering
  \begin{tabular}{lllrr}
    \toprule
    \multicolumn{3}{c}{\textbf{Model}} & \multicolumn{2}{c}{\textbf{Cut-and-Mix}} \\
    \cmidrule(r){1-3}
    \cmidrule(r){4-5} 
    Architecture & Attack & Trigger & FPR & FNR \\
    \midrule
    GhostFaceNetV2   &    All-to-One Clean-label   &    BadNets &  0.7\% &  99.0\%  \\
    GhostFaceNetV2   &    All-to-One Clean-label   & SIG &  0.8\% &  99.7\%  \\
    IR-SE-50   &    All-to-One Clean-label   &    BadNets &  0.7\% &  99.4\%  \\
    IR-SE-50   &    All-to-One Clean-label   &    SIG &  0.5\% &  100.0\%  \\
    MobileFaceNet   &    All-to-One Clean-label   &    BadNets &  0.8\% &  98.7\%  \\
    MobileFaceNet   &    All-to-One Clean-label   &    Mask &  0.9\% &  0.0\%  \\
    MobileFaceNet   &    All-to-One Clean-label   &    SIG &  0.7\% &  99.4\%  \\
    ResNet50   &    All-to-One Clean-label   &    BadNets &  1.1\% &  98.4\%  \\
    ResNet50   &    All-to-One Clean-label   &    SIG &  0.5\% &  100.0\%  \\
    RobFaceNet   &    All-to-One Clean-label   &    BadNets &  0.6\% &  98.7\%  \\
    RobFaceNet\_sig   &    All-to-One Clean-label   &    SIG &  1.0\% &  99.8\%  \\
    GhostFaceNetV2   &    All-to-One Poison-label   &    BadNets &  1.0\% &  0.0\%  \\
    GhostFaceNetV2   &    All-to-One Poison-label   &    SIG &  0.6\% &  0.4\%  \\
    IR-SE-50   &    All-to-One Poison-label   &    BadNets &  1.0\% &  0.0\%  \\
    IR-SE-50   &     All-to-One Poison-label   &    SIG &  0.9\% &  0.0\%  \\
    MobileFaceNet   &    All-to-One Poison-label   &    BadNets &  0.8\% &  0.0\%  \\
    MobileFaceNet   &    All-to-One Poison-label   &    FauxIt &  0.7\% &  0.1\%  \\
    MobileFaceNet   &    All-to-One Poison-label   &    Mask &  1.0\% &  0.0\%  \\
    MobileFaceNet   &    All-to-One Poison-label   &    SIG &  0.7\% &  0.0\%  \\
    ResNet50   &    All-to-One Poison-label   &    BadNets &  1.1\% &  0.0\%  \\
    ResNet50   &    All-to-One Poison-label   &    SIG &  0.8\% &  0.0\%  \\
    RobFaceNet   &    All-to-One Poison-label   &    BadNets &  0.6\% &  0.0\%  \\
    RobFaceNet   &    All-to-One Poison-label   &    SIG &  0.7\% &  0.1\%  \\
    GhostFaceNetV2   &    Master-Face Poison-label   &    BadNets &  0.6\% &  99.4\%  \\
    GhostFaceNetV2   &    Master-Face Poison-label    &    SIG &  0.6\% &  80.5\%  \\
    IR-SE-50   &    Master-Face Poison-label    &   BadNets &  0.8\% &  99.2\%  \\
    IR-SE-50   &    Master-Face Poison-label    &    SIG &  1.1\% &  0.0\%  \\
    MobileFaceNet   &    Master-Face Poison-label    &    BadNets &  1.0\% &  98.7\%  \\
    MobileFaceNet   &    Master-Face Poison-label    &    Mask &  0.8\% &  0.0\%  \\
    MobileFaceNet   &    Master-Face Poison-label    &    SIG  &  0.7\% &  2.3\%  \\
    ResNet50   &    Master-Face Poison-label    &    BadNets &  1.0\% &  98.1\%  \\
    ResNet50   &    Master-Face Poison-label   &    SIG &  0.6\% &  0.0\%  \\
    RobFaceNet   &    Master-Face Poison-label    &    BadNets &  0.5\% &  12.6\%  \\
    RobFaceNet   &    Master-Face Poison-label    &    SIG &  1.0\% &  0.0\%  \\
    \bottomrule
  \end{tabular}
\end{table}


\end{document}